# Resolución geométrica tridimensional de la cinemática inversa de un brazo articulado de 7 grados de libertad

Cinemática de sistemas robóticos

# Three-dimensional geometric resolution of the inverse kinematics of a 7 degree of freedom articulated arm

Kinematics of robotic systems

Dr. Antonio Losada González
anlosada@uvigo.es
Universidad de Vigo

## 1. Abstract

This work presents a three-dimensional geometric resolution method to calculate the complete inverse kinematics of a 7-degree-of-freedom articulated arm, including the hand itself. The method is classified as an analytical method with geometric solution, since it obtains a precise solution in a closed number of steps, converting the inverse kinematic problem into a three-dimensional geometric model. To simplify the problem, the kinematic decoupling method is used, so that the position of the wrist is calculated independently on one hand with information on the orientation of the hand, and the angles of the rest of the arm are calculated from the wrist.

## 2. Resumen

Este trabajo presenta un método de resolución geométrica en tres dimensiones para calcular la cinemática inversa completa de un brazo articulado de 7 grados de libertad contando la propia mano. El método se encuadra dentro de los métodos analíticas con solución geométrica, dado que obtiene una solución precisa en un número de pasos cerrado convirtiendo el problema cinemático inverso en un modelo geométrico en tres dimensiones. Para simplificar el problema se emplea el método del desacoplo cinemático, de forma que se calcula de forma independiente la posición de la muñeca por una lado con información de la orientación de la mano, y a partir de la muñeca se calculan los ángulos del resto del brazo.

## 3. Introducción

La resolución cinemática inversa puede realizarse por métodos analíticos o numéricos. Los segundos son iterativos y consumen gran cantidad de potencia de cómputo por lo que resultan preferibles los métodos analíticos, y dentro de estos últimos tenemos los métodos algebraicos y los métodos geométricos. El método propuesto en este trabaja se encuentra dentro de los métodos analíticos con resolución geométrica en tres dimensiones. Trabajos similares podemos encontrarlos en [2][16] en donde se proponen soluciones geométricas en tres dimensiones. Como métodos iterativos numéricos tenemos trabajos como [9][12]. En el caso de que el robot tenga un grado menos de libertad, los métodos geométricos pueden reducir la complejidad a un sistema geométrico de dos dimensiones, lo que reduce en gran medida su complejidad, ejemplos de métodos de resolución geométrica 2D los encontramos en trabajos como [3][5][8][10]. Otros trabajos emplean métodos numéricos, pero para reducir la complejidad del espacio de posibles soluciones emplean o lógica difusa como [6][13] o redes neuronales como [7]. En gran parte de los trabajos se expone la resolución de la cinemática directa como [4][11][15], que resulta importante para conocer las necesidades de resolución de la cinemática inversa.

## 4. Descripción del robot

En este trabajo pretendemos introducir un modo muy poco común de calcular la cinemática inversa de un brazo de 6 grados de libertad, siendo posible llegar con esta técnica a 7 grados de libertad, sin contar con el movimiento de apertura y cierre de la pinza. Debemos tener en cuenta que en este tipo de brazo, para el cálculo de la cinemática inversa se emplea la técnica del desacoplo cinemático, por lo que el problema se divide en dos partes, en una se resuelve el punto donde debe encontrarse la muñeca y en la otra se resuelven los ángulos del brazo y antebrazo. En este caso, lo que hace que este problema sea especialmente complejo es la existencia de 4 grados de libertad en el brazo y antebrazo.

Normalmente los brazos articulados de 5 grados de libertad o menos con 3 grados de libertad o menos en el brazo y antebrazo, se resuelven mediante sistemas geométricos en 2D empleando el teorema del coseno, en nuestro caso debemos resolverlo mediante figuras geométricas en 3D, siendo muy extraño encontrar trabajos en los que se proponga este tipo de resolución geométrica.

El objetivo es aplicar los resultados de este trabajo a un robot real. Hemos escogido el robot meccanoid gs15k debido a su disponibilidad y a su bajo precio. Este robot puede controlarse mediante bluetooth empleando Python mediante el código que se encuentra en la siguiente página https://sites.google.com/view/inteligenciarobotica/inicio. Sus brazos, inicialmente solo tienen 3 grados de libertad, por lo que ha sido necesario realizar una serie de modificaciones para alcanzar los 7 grados. Se le ha incorporado un servo de meccano en cada brazo (puede verse en color azul en la Imagen 1 y adicionalmente se han incorporado 3 servos estándar más en cada una de las manos que permiten dos tipo de giro de muñeca y la apertura y cierre de la mano. Adicionalmente ha sido necesario realizar ciertas modificaciones en la mano para permitir que esta sea articulada (las modificaciones están detalladas en https://sites.google.com/view/inteligenciarobotica/inicio).

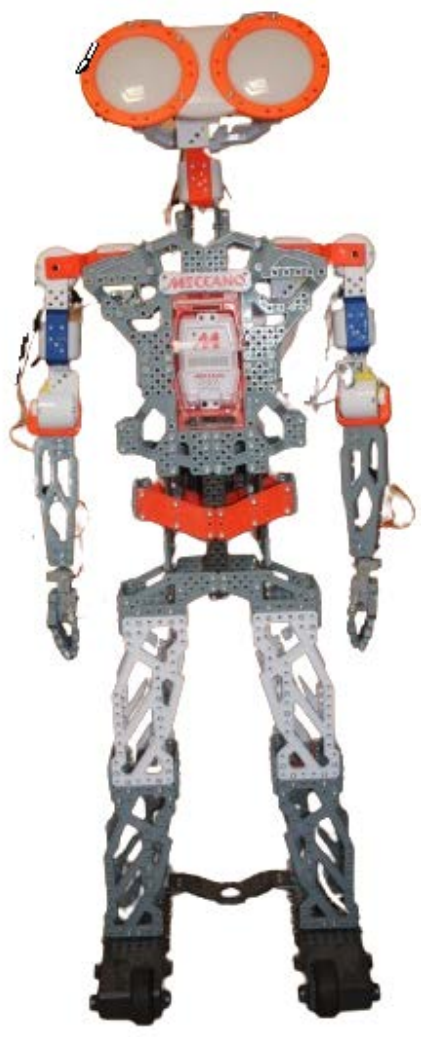

*Imagen 1. Robot meccanoid G15KS*

Los servos del propio robot pueden controlarse mediante el protocolo propio de meccano empleando bluetooth sin electrónica adicional, pero los nuevos servos deben ser controlados mediante hardware adicional, para lo que ha sido necesario desarrollar una placa electrónica en la que montamos un arduino mini pro que será controlado desde un ordenador enviando los comandos mediante bluetooth en una primera versión. El arduino puede ser substituido por un equipo más potente como una raspberry o un módulo Jetson  de Nvidea en el caso de que se quiera controlar directamente el robot mediante aplicaciones de inteligencia artificial empotradas. En nuestro caso, el arduino controlará los dos servos adicionales de meccano junto con los 6 servos estándar de las manos.

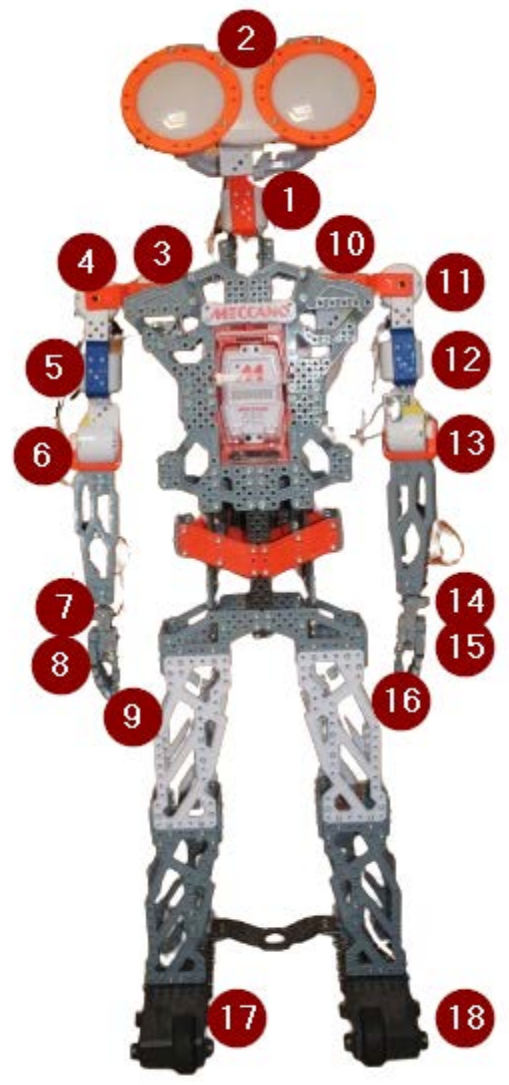

*Imagen 2. Componentes de meccanoid G15KS*

En la Imagen 2 podemos ver el robot junto con la enumeración de todos sus componentes que pasamos a describir:

1. Cuello articulado con dos grados de libertad permitiendo el movimiento de cabeceo y guiñada.

2. Cabeza con iluminación led en color. En la versión final contará con visión estereoscópica 3D y con un panel de leds para mostrar la expresión de la boca.
3. Servo del hombro derecho que permite un grado de libertad
4. Servo del hombro derecho que permite un grado de libertad
5. Servo adicional añadido a la configuración inicial que permite un giro del antebrazo
6. Servo del codo derecho
7. Servo de la muñeca derecha
8. Servo que permite girar la mano derecha
9. Servo que permite la apertura y cierre de la mano derecha
10. Servo del hombro izquierdo que permite un grado de libertad
11. Servo del hombro izquierdo que permite un grado de libertad
12. Servo adicional añadido a la configuración inicial que permite un giro del antebrazo
13. Servo del codo izquierdo
14. Servo de la muñeca izquierda
15. Servo que permite girar la mano izquierda
16. Servo que permite la apertura y cierre de la mano izquierda
17. Motor del pie derecho
18. Motor del pie izquierdo

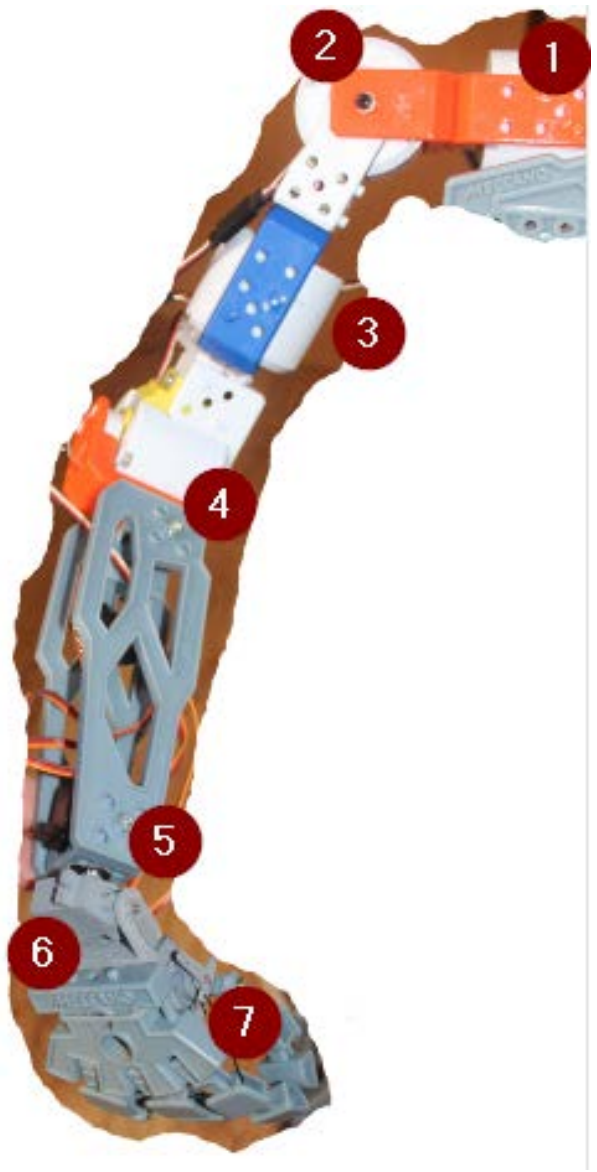

*Imagen 3. Componentes del brazo del robot*

En la Imagen 3 se muestra cada uno de los servos que permiten el movimiento del brazo. Cada servo permite un grado de libertad independiente. El servo 7 es particular ya que permite la apertura y cierre de la mano. Para poder calcular la cinemática inversa de un brazo de 7 grados de libertad como el que se muestra en la Imagen 4 empleando métodos geométricos emplearemos varias técnicas. Por una lado comenzaremos por emplear la técnica del desacoplo cinemático, de forma que dividimos el brazo completo en dos conjuntos de articulaciones, por un lado tenemos el hombro y el codo con 4 grados de libertad compuesto por los servos 1, 2, 3 y 4 y por otro lado tenemos la muñeca y la mano con tres grados de libertad compuesto por los servos 5, 6, y 7. Para ubicar de forma precisa el eyector final en un punto concreto necesitamos conocer tanto el punto final del extremo de la articulación (identificado por un (1) en la Imagen 4) como la orientación del vector que une la unión de la muñeca (2) con el extremo del eyector final (1), como se muestra en la Imagen 4.

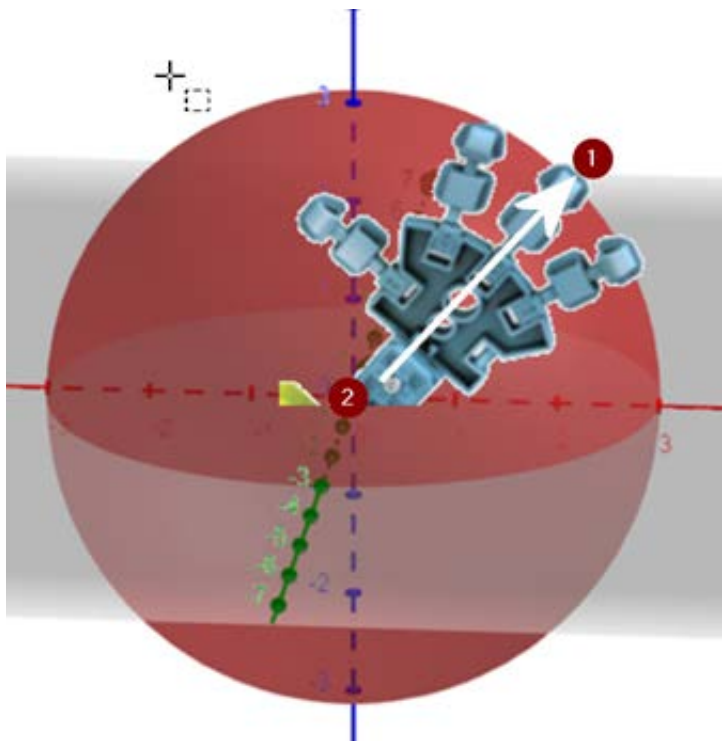

*Imagen 4. Esfera virtual que puede trazar el extremo de la mano*

En la Imagen 4 podemos ver la esfera que representa los puntos finales de la mano, asumiendo que la mano tiene una longitud de 3 unidades. En la Imagen 4 se muestra el punto (1) que indica el punto final en el que debe posicionarse el extremo de la mano. Conociendo la orientación de la mano con respecto al punto (1) debemos calcular el punto (2) que nos indica la posición de la muñeca. Conociendo este punto podemos iniciar los cálculos necesarios para posicionar el resto del brazo.

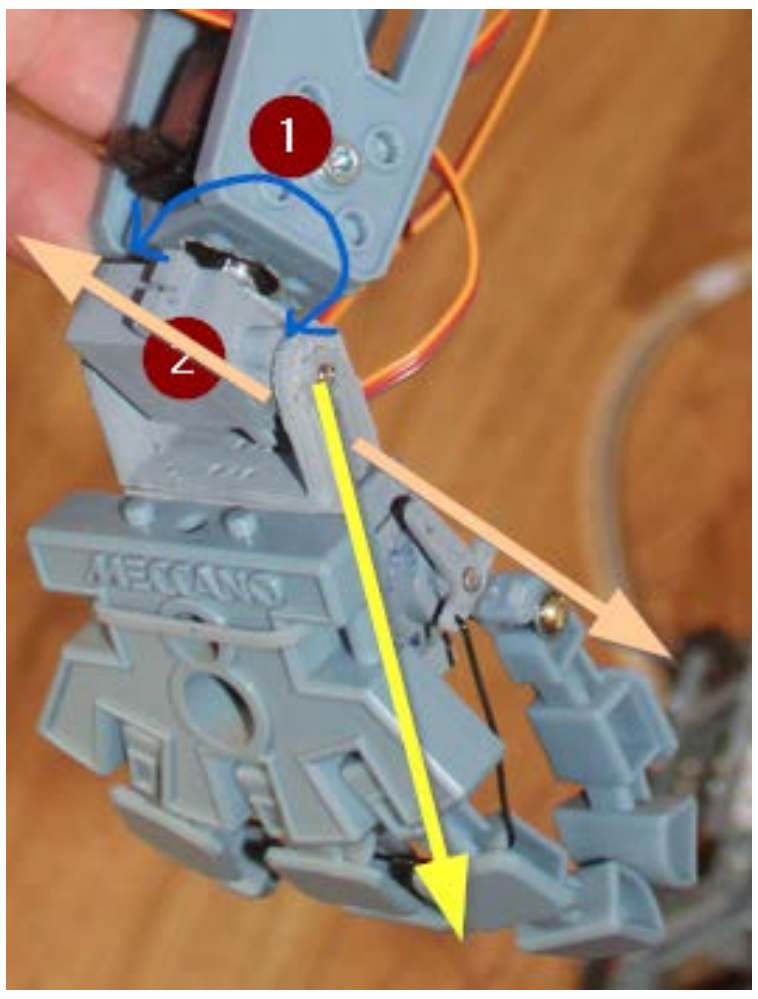

*Imagen 5. Movimientos de la muñeca*

En la Imagen 5 podemos apreciar los servos que intervienen en cada uno de los grados de libertad. En la Imagen 5 se señalan los servos de la muñeca (1) y de la mano (2). El servo de la muñeca permite un giro rotacional posible de 360 grados que se encuentra limitado por el mecanismo del servo a 180 grados. Las direcciones del movimiento se marcan con la trayectoria en azul y el servo de la mano (2) permite moverla hacia arriba y abajo como indican los vectores naranjas. La longitud del vector amarillo indica la longitud de la mano y geométricamente representará el radio de la esfera de la Imagen 4.

## 5. Estudio geométrico

Si no hubiera limitaciones físicas, el primer grado de libertad (1) permitiría un giro de 360 grados y el segundo permitiría 180 grados. Realmente, todos los servos estándar están limitados a 180° a lo que tenemos que añadir las limitaciones mecánicas debido a la posición concreta de las articulaciones.

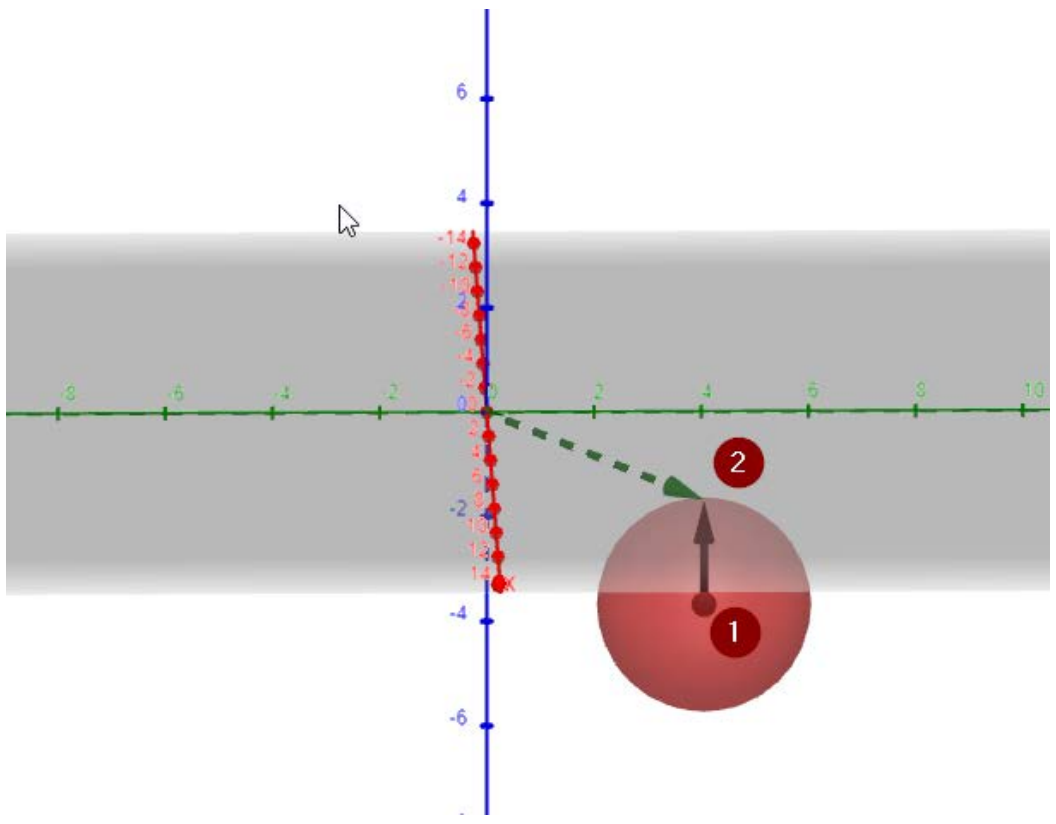

*Imagen 6. Representación geométrica 3D de la muñeca*

Trasladando las articulaciones de la Imagen 5 a un esquema geométrico en 3D como el que se muestra en la Imagen 6, podemos ver que la punta de la mano (2) podría alcanzar todos los puntos de una hipotética esfera con centro en (1), si asumimos que el punto (1) es el extremo de la mano y el punto (2) es el punto que marca la muñeca (pueden verse representados en la Imagen 4). Conociendo el punto (1), para calcular el punto (2) solo necesitamos conocer el ángulo de la mano, el ángulo de giro de la muñeca y la longitud de la mano.

Con estos datos iniciales, aplicando las ecuaciones (1)-(10) llegamos a obtener el punto donde debe situarse la muñeca (2) ($P_{muneca}$). Este será nuestro punto de partida para calcular los ángulos del resto del brazo.

$$ang_{muneca} = 0 \tag{1}$$

$$ang_{mano} = 0 \tag{2}$$

$$long_{mano} = 2 \tag{3}$$

$$P_x = 3 \tag{4}$$

$$P_y = 4 \tag{5}$$

$$P_z = -3 \tag{6}$$

$$M_x = P_x + long_{mano} \cdot sen(ang_{muneca}) \cdot \cos(ang_{mano}) \tag{7}$$

$$M_y = P_y + long_{mano} \cdot sen(ang_{muneca}) \cdot sen(ang_{mano}) \tag{8}$$

$$M_z = P_z + long_{mano} \cdot cos(ang_{muneca}) \tag{9}$$

$$P_{muneca} = (M_x, M_y, M_z) \tag{10}$$

$$eq.esfera\colon (x - P_x)^2 + (y - P_y)^2 + (z - P_z)^2 = {long_{mano}}^2 \tag{11}$$

Para obtener las coordenadas del punto (2) solo debemos aplicar las ecuaciones paramétricas de la esfera. Para resolver la posición en el espacio de $P_{muneca}$ necesitamos conocer los ángulos $ang_{muneca}$ y $ang_{mano}$. El primero está representado en la Imagen 7 como el ángulo (1) pudiendo tener valores entre 0 y π y el en segundo está representado por el ángulo (4) pudiendo tener valores entre 0 y π/2.

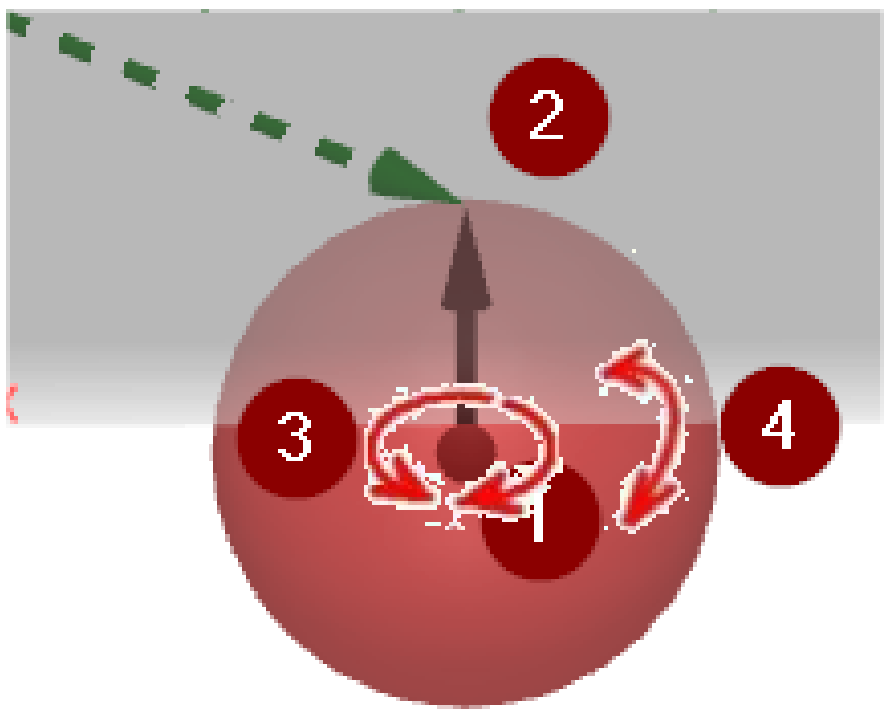

*Imagen 7. Ángulos de giro de la mano*

En la Imagen 7 podemos ver gráficamente reflejados los ángulos $ang_{muneca}(1)$ y $ang_{mano}(4)$, el primero de ellos es rotacional de 360 grados permitiendo un giro completo sobre el plano horizontal y el segundo es rotacional de 180 grados permitiendo medio giro hacia arriba y abajo en el plano vertical.

Debemos tener en cuenta que la posición de la muñeca calculada (2) es un punto ideal, que puede ser posible que no se pueda alcanzar por limitaciones físicas del robot, por lo que una vez finalizado el cálculo de toda la cinemática inversa, debemos comprobar si los ángulos son posibles y en caso contrario los reajustaremos de nuevo.

En la segunda parte del cálculo, conociendo el punto en el que debe situarse la muñeca, debemos calcular el resto de los ángulos que deben alcanzar los servos que permiten los ángulos de libertar del brazo, antebrazo y hombro.

Una vez obtenido el punto (2) de la Imagen 7 tenemos la posición de la muñeca, con lo que estamos en disposición de calcular el resto de los puntos importantes en el brazo.

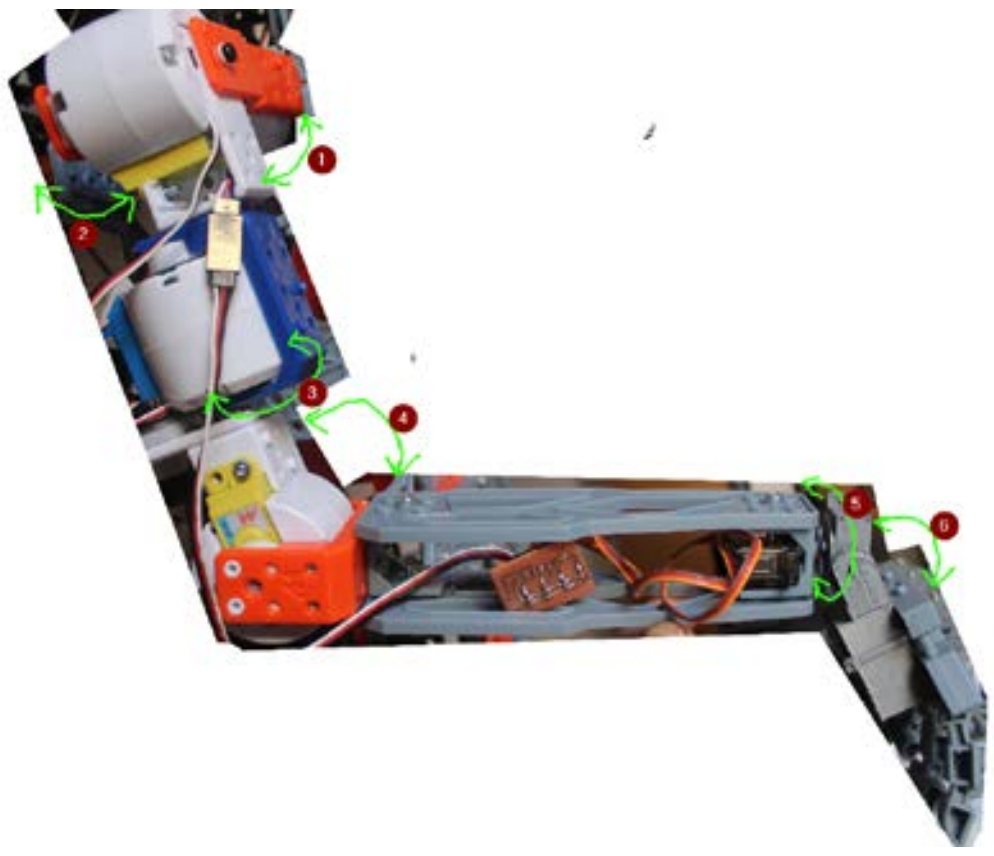

*Imagen 8. Visión lateral de los componentes del brazo del robot*

En la Imagen 8 se pueden ver con claridad todos los servos que componen el brazo del robot junto con los movimientos que permiten cada uno de ellos marcados en color verde.

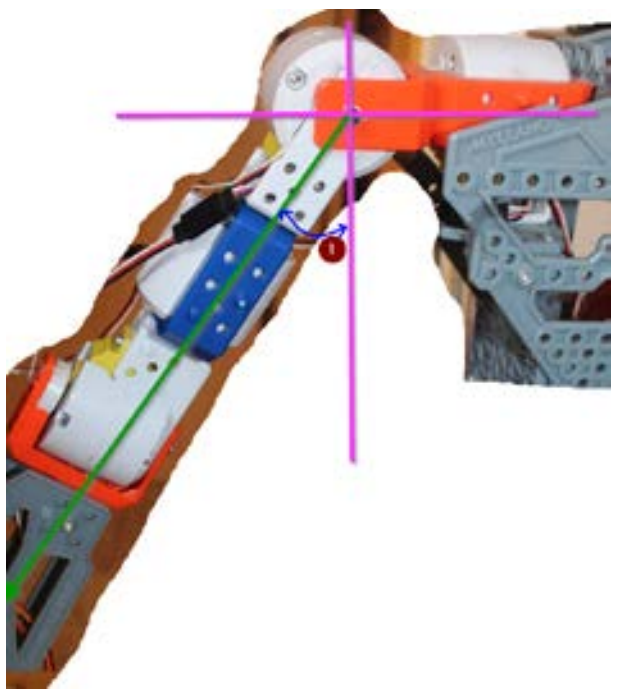

*Imagen 9. Visión frontal del ángulo del hombro*

Para facilitar los cálculos, en la Imagen 9, el centro de la cruz marcada por la intersección de los dos ejes de coordenadas morados será nuestro origen de coordenadas espaciales, desde este punto se calcularán las posiciones del resto de los puntos que serán necesarios para resolver la cinemática inversa del robot.

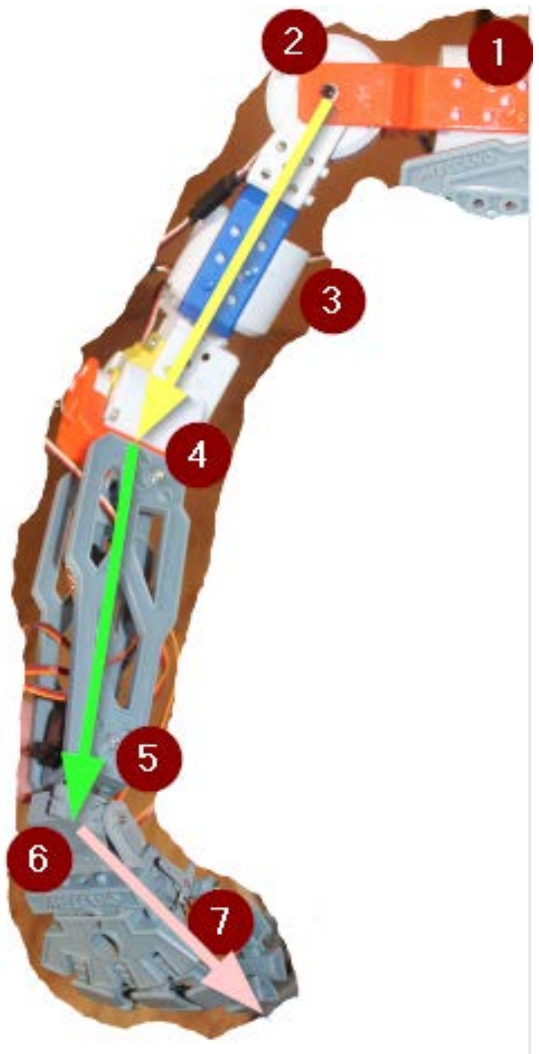

*Imagen 10. Vectores virtuales que componen el brazo*

Para alcanzar la representación geométrica del brazo robótica, se dividirá en segmentos, de forma que cada segmento esté representado por un vector. El segmento amarillo es el brazo, el segmento verde es el antebrazo y el segmento rosa es la mano. En la Imagen 11 podemos visualizar la conversión de todos los segmentos de la Imagen 10 a un modelo geométrico tridimensional.

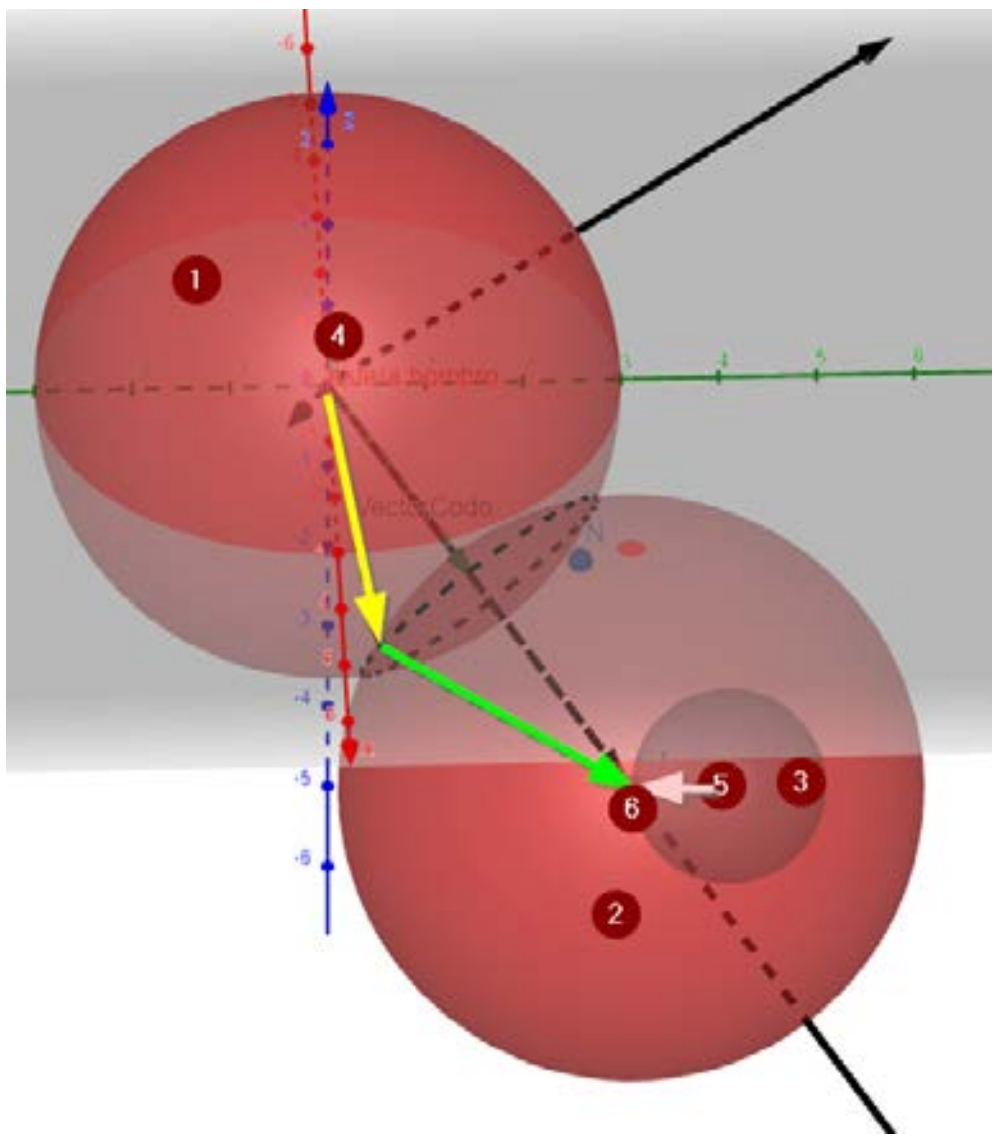

*Imagen 11. Representación geométrica de los vectores del brazo y la mano*

En la Imagen 11 podemos ver los siguientes objetos:

1. Esfera virtual generada por el borde del brazo (codo) con centro en el hombro (punto 4)
2. Esfera virtual generada por el origen del antebrazo (codo) con centro en la muñeca (punto 6)
3. Esfera virtual generada por la posición de la muñeca con centro en el punto más alejado del efector final o mano (punto 5)
4. Posición del hombro
5. Posición del borde la mano
6. Posición de la muñeca

Puede apreciarse como se mantienen los colores de los vectores o segmentos entre la Imagen 10 y la Imagen 11.

En nuestra representación denominamos codo a la unión entre el brazo y el antebrazo. Si trasladamos estos segmentos a una representación geométrica en el espacio tridimensional podemos apreciar que si no hubiera limitaciones físicas, el borde del segmento del brazo podría dibujar una esfera con centro en el hombro (4) y en cuya superficie tiene que encontrarse el punto que define el codo y de igual forma, visto desde la muñeca (6), el antebrazo define otra esfera con centro en la muñeca (6) y en cuya superficie tiene que encontrarse el codo (enlace de los vectores amarillo y verde). En la Imagen 11 se puede ver de forma clara que el codo debe encontrarse siempre comprendido dentro de la circunferencia de intersección de las esferas que tienen como centro el hombro (4) y la esfera que tiene como centro la muñeca (6). Si resolvemos las ecuaciones que nos permiten obtener los puntos de la circunferencia de intersección, tendremos todos los puntos válidos para ubicar el codo, y dado que conocemos los puntos (4) y (6) podemos resolver todo el sistema. Una vez obtenidas las ecuaciones paramétricas de la circunferencia de intersección, cualquiera de sus puntos sería válido como resolución cinemática del problema, pero en este punto entrarán las limitaciones físicas del robot o la trayectoria con menos gasto de energía desde la posición actual para delimitar las soluciones más óptimas.

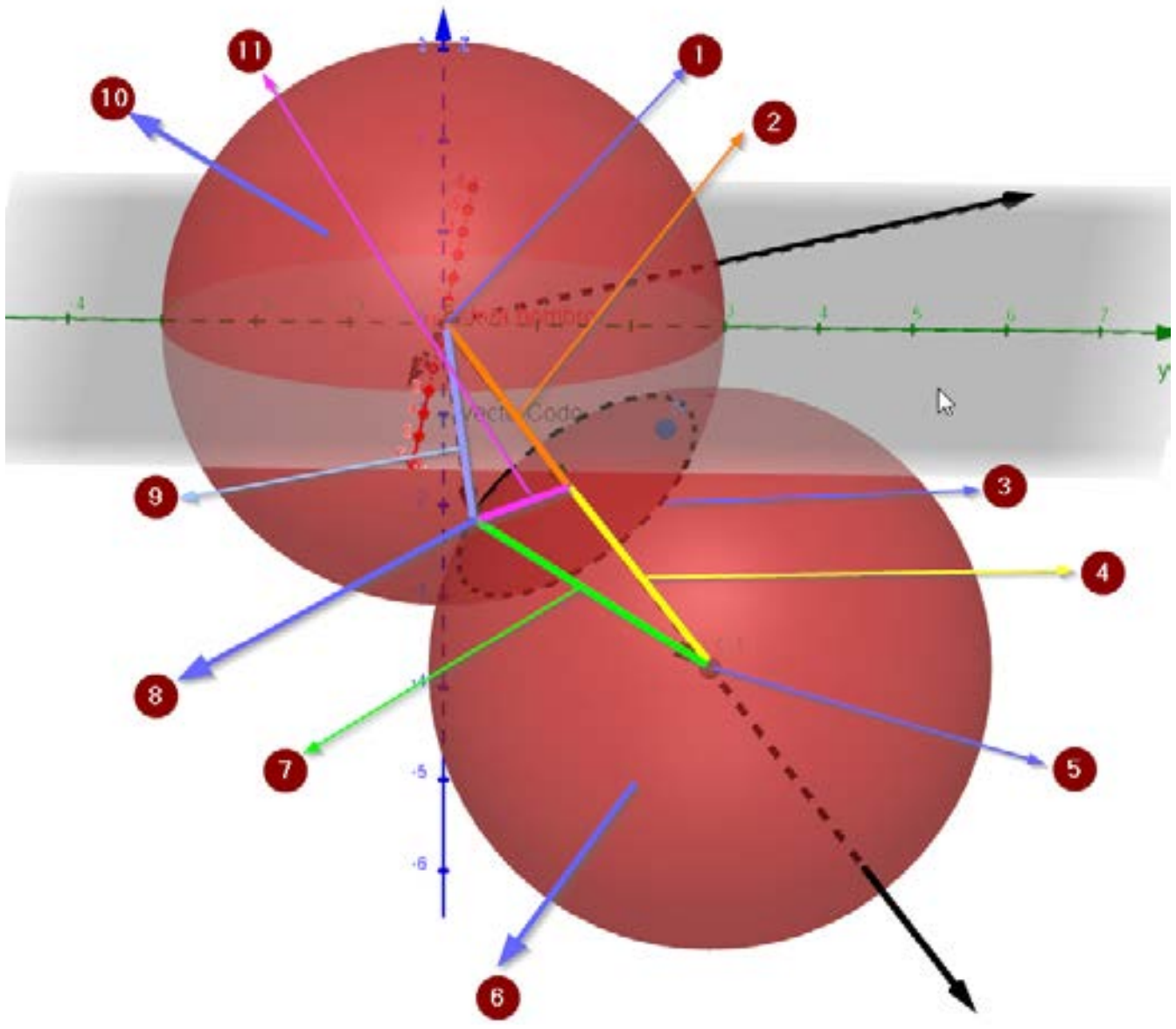

*Imagen 12. Componentes de la representación geométrica del brazo*

Empleando el desacoplo cinemático se ha calculado el punto (6) de la Imagen 11, con lo que queda por calcular la posición y orientación del brazo y antebrazo. En la Imagen 12 se pueden observar todos los objetos que entrarán en juego para resolver la cinemática inversa del resto del brazo.

1. Localización del hombro del robot. Será el punto central de intersección entre los ejes de coordenadas de nuestro espacio tridimensional. Es el centro de la esfera que podría generar el borde del segmento del brazo (codo) si no tuviera limitación de movimiento.
2. Parte del vector que une el hombro con la posición final de la muñeca. La primera parte une el hombro con el centro de la circunferencia de intersección entre las esferas de movimiento del brazo y el antebrazo (naranja).
3. Circunferencia punteada de intersección entre las dos esferas que representan los movimientos posibles del brazo y del antebrazo. Este es el lugar donde debe encontrarse el codo. Cualquier punto de la circunferencia es válido para alcanzar el punto final calculado de la muñeca (5)
4. Parte del vector que une el hombro con la posición final de la muñeca. La segunda parte une el centro de la circunferencia de intersección entre las esferas de movimiento del brazo y el antebrazo con el punto final de la muñeca (amarillo).
5. Punto en donde debe situarse la muñeca.
6. Esfera teórica que podría alcanzar el antebrazo si no tuviera limitaciones en su movimiento
7. Vector que representa el antebrazo
8. Representa el codo que es el punto de unión entre el brazo y el antebrazo
9. Vector que representa el brazo
10. esfera que podría generar el borde del segmento del brazo si no tuviera limitación de movimiento.
11. Vector perpendicular al segmento que une el hombro con la muñeca pasando por el codo. Representado en las fórmulas como ***catCodo***, dado que es un cateto de los dos triángulos rectángulos formados por el brazo y por el antebrazo. En la imagen no parecen triángulos rectángulos por la perspectiva.

Una vez identificados todos los elementos que debemos tener en cuenta en el sistema, el objetivo es obtener las ecuaciones paramétricas de la circunferencia de intersección entre las dos esferas. Una vez obtenidas, debemos escoger algún punto físicamente válido dentro de la circunferencia para resolver la posición del codo y una vez resuelta la posición del codo, debemos calcular los ángulos del segmento que define el brazo y el antebrazo. Conociendo los ángulos de estos segmentos debemos recalcular los ángulos de la muñeca y la mano, dado que los ángulos conocidos hasta este punto son los ángulos proporcionados desde el punto final de la mano, pero para posicionar los servos de la muñeca necesitamos los ángulos desde el punto de vista de la muñeca.

Como se ha empleado el desacoplo cinemático, hemos separado los cálculos de la muñeca y los del resto del brazo. Hasta este punto se ha calculado la posición de la muñeca con la posición del extremo de la mano y los ángulos de la mano. Todos los cálculos a partir de este punto se basan en el conocimiento del punto de la muñeca. En la ecuación (12) se asigna un punto de la muñeca para proceder a realizar los cálculos del resto del artículo con valores sencillos para simplificar los cálculos.

$$a = M_x = 3, b = M_y = 3, c = M_z = -3 \quad (12)$$

$$C1 = (a, b, c) = (3,3,-3) \rightarrow Punto\ 5, Imagen\ 12 \quad (13)$$

$$d_1 = 3 \rightarrow longitud\ del\ brazo \quad (14)$$

$$d_2 = 3 \rightarrow longitud\ del\ antebrazo \quad (15)$$

$$Esfera\ Brazo:\ x^2 + y^2 + z^2 = {d_1}^2 \rightarrow Centro\ (0,0,0) \quad (16)$$

$$Esfera\ antebrazo: (x-a)^2 + (y-b)^2 + (z-c)^2 = {d_2}^2 \rightarrow Centro\ (a, b, c) \quad (17)$$

La Ecuación 16 representa la esfera marcado con un 10 en la Imagen 12, mientras la Ecuación (17) representa la esfera marcada con un 6. La primera se encuentra en el origen de coordenadas (0,0,0), mientras que la segunda tiene como centro el punto $(a, b, c)$.

Para poder obtener las ecuaciones paramétricas de la circunferencia de intersección entre las dos esferas, primero debemos obtener el plano de intersección que contendrá todos los puntos de la circunferencia. Este plano puede observarse en la Imagen 13 y está definido por la Ecuación (18).

Para obtener la ecuación del plano (18), resolvemos el sistema de ecuaciones compuesto por la Ecuación (16) y la Ecuación (17).

$$Ecuación\ del\ plano = 2ax + 2by + 2cz - a^2 - b^2 - c^2 = {d_1}^2 - {d_2}^2 = \quad (18)$$

$$Ecuación\ del\ plano: 2x + 2y - 2z = 9 \quad (19)$$

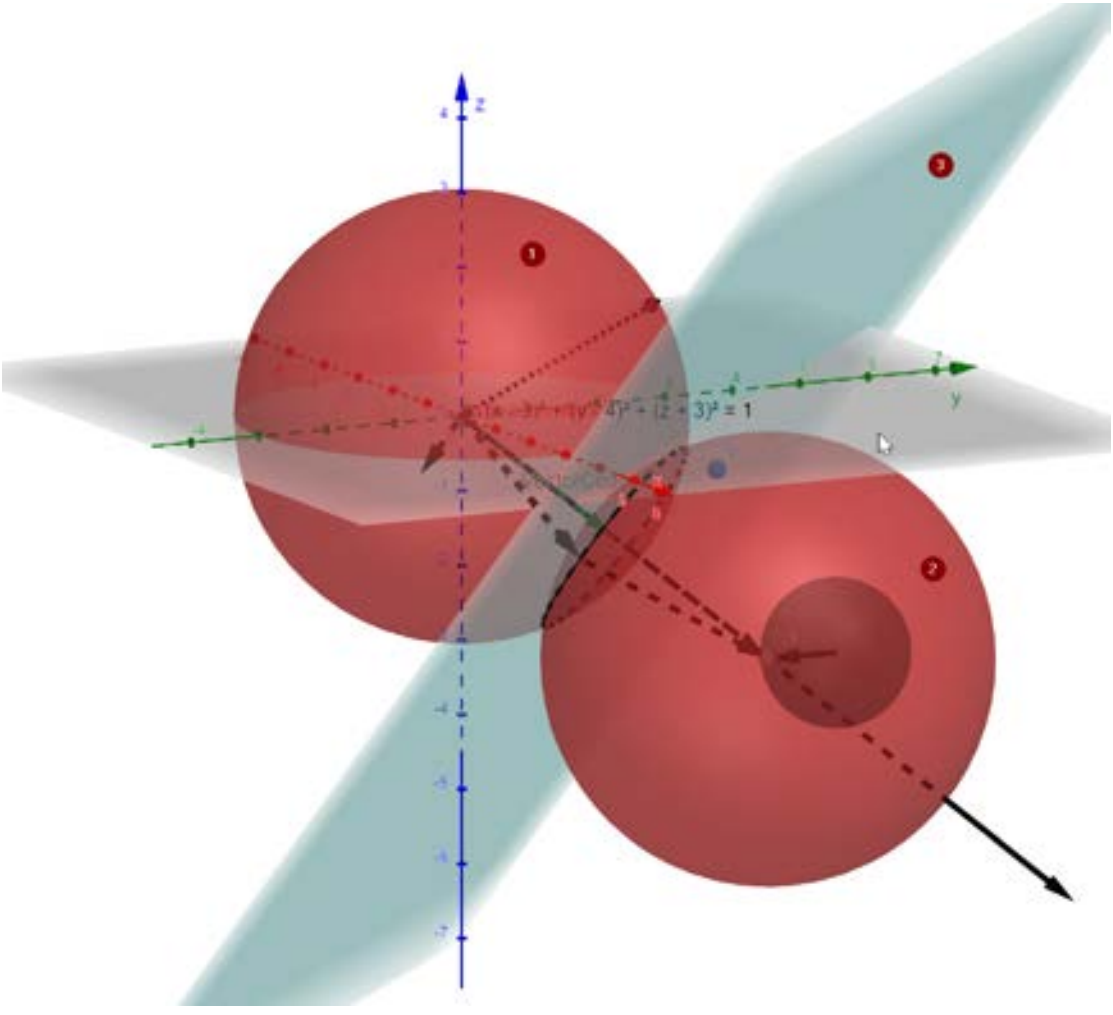

*Imagen 13. Plano de intersección entre las dos esferas.*

Una vez calculado el plano de intersección, debemos generar un nuevo conjunto de ejes de coordenadas que nos permitan pintar la circunferencia de intersección encima del plano de intersección de las dos esferas. Para ello comenzamos por generar un vector perpendicular al plano que debe partir del eje de coordenadas actual (punto 1 de la imagen 12) y atravesar el punto que define la posición de la muñeca. Este vector debe pasar por el punto $(a, b, c)$ (punto 5 de la imagen 12) y se denominará $\overrightarrow{v_1}$ como indica la Ecuación $(20)$.

Para generar otro de los ejes de coordenadas, intercambiamos las coordenadas $x, y$ del vector $v_1$ de Ecuación $(20)$ e invertimos el signo de la coordenada $x$, dejando la coordenada $z$ como 0, obteniendo el vector $\overrightarrow{a_1}$ de la Ecuación $(21)$. Este vector es perpendicular al vector $v_1$ encontrándose en su mismo plano.

Finalmente para obtener el tercer vector realizaremos un producto vectorial de los dos vectores $a_1$ y $v_1$, dado que este producto genera un vector perpendicular a ambos, con lo que obtenemos el vector $\overrightarrow{b_1}$ de la Ecuación $(22)$.

$$\overrightarrow{v_1} = \begin{pmatrix} 2a \\ 2b \\ 2c \end{pmatrix} = \begin{pmatrix} 6 \\ 6 \\ -6 \end{pmatrix} \qquad (20)$$

$$\overrightarrow{a_1} = \begin{pmatrix} -2b \\ 2a \\ 0 \end{pmatrix} = \begin{pmatrix} -6 \\ 6 \\ 0 \end{pmatrix} \qquad (21)$$

$$\overrightarrow{b_1} = a_1 \otimes v_1 = \begin{pmatrix} -36 \\ -36 \\ -72 \end{pmatrix} \qquad (22)$$

Los vectores $\overrightarrow{a_1}$ y $\overrightarrow{b_1}$ serán los que definan los ejes de coordenadas $x, y$ de nuestro plano de intersección de esferas.

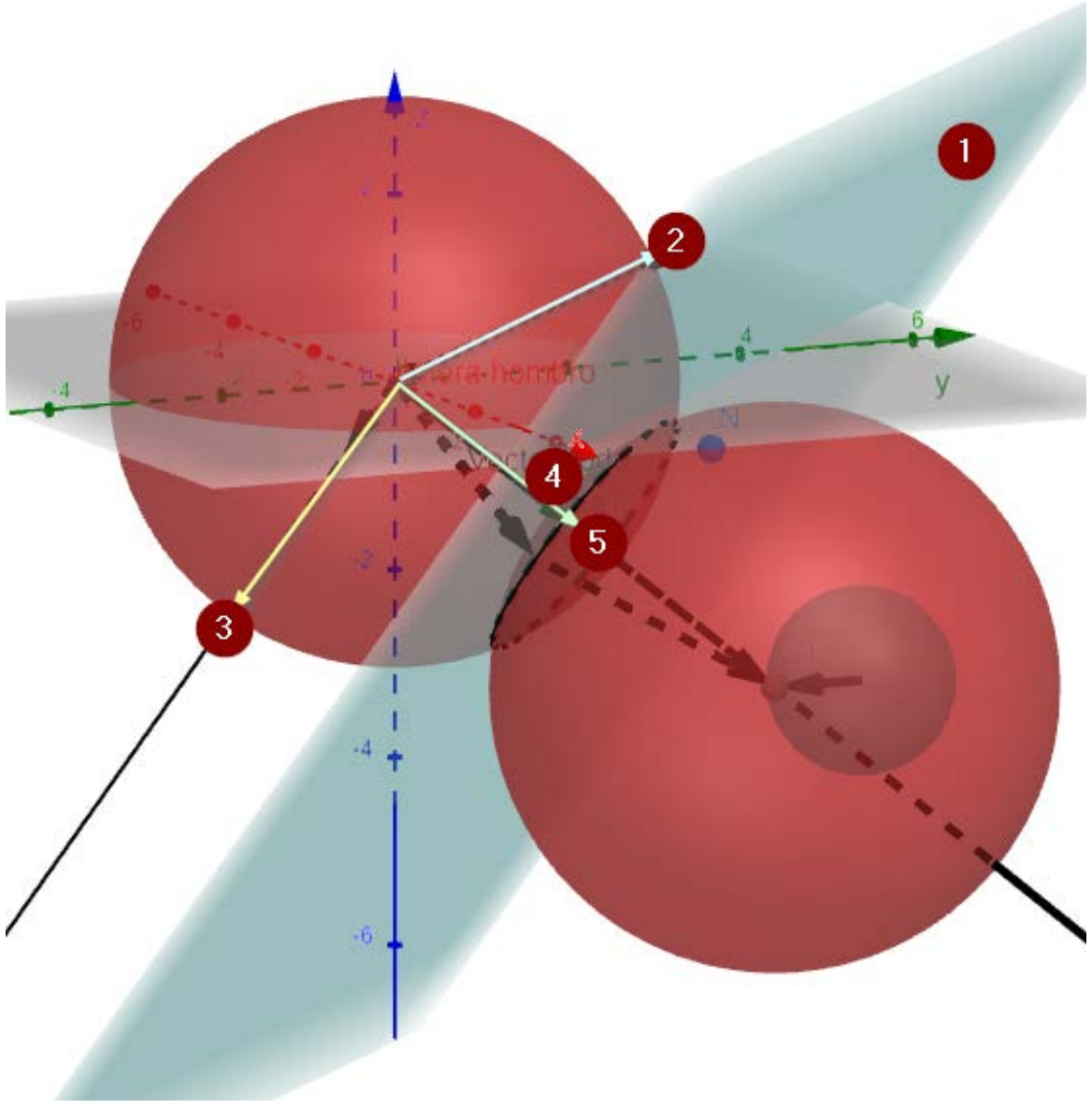

*Imagen 14. Ejes de coordenadas del plano de intersección.*

En la imagen 14 se visualizan los vectores $\overrightarrow{v_1}$ (5), $\overrightarrow{a_1}\,(2)$ y $\overrightarrow{b_1}(3)$. Los vectores identificados como (2), (3) definen un plano paralelo al plano (1). El siguiente paso será normalizar los vectores $\overrightarrow{a_1}\,(2)$ y $\overrightarrow{b_1}(3)$ y calcular el centro de la circunferencia de intersección (5).

El siguiente paso será la normalización de estos vectores comenzando por el vector $\overrightarrow{a_1}$, para ello calculamos el módulo (Ecuación $(23)$ y $(24)$) y posteriormente se divide cada una de sus componentes por dicho módulo (Ecuación $(25)$)

$$\|a_1\| = \sqrt{\sum_{i=1}^{n}(x_i)^2} \qquad (23)$$

$$\|a_1\| = \sqrt{(-2b)^2 + (2a)^2} = 8.49 \qquad (24)$$

$$\overrightarrow{an_1} = \frac{\overrightarrow{a_1}}{\|a_1\|} = \begin{pmatrix} \frac{-2b}{\|a_1\|} \\ \frac{2a}{\|a_1\|} \\ \frac{0}{\|a_1\|} \end{pmatrix} = \begin{pmatrix} -0.71 \\ 0.71 \\ 0 \end{pmatrix} \qquad (25)$$

Normalizamos el vector $\overrightarrow{b_1}$:

$$\|b_1\| = \sqrt{(-36)^2 + (-36)^2 + (-72)^2} \qquad (26)$$

$$\overrightarrow{bn_1} = \frac{\overrightarrow{b_1}}{\|b_1\|} = \begin{pmatrix} \frac{-36}{\|b_1\|} \\ \frac{-36}{\|b_1\|} \\ \frac{-72}{\|b_1\|} \end{pmatrix} = \begin{pmatrix} -0.41 \\ -0.41 \\ -0.82 \end{pmatrix} \quad (27)$$

En este punto tenemos definidos los vectores que nos servirán de eje de coordenadas para trazar la circunferencia de intersección en el plano, con lo que calcularemos el centro y el radio dela circunferencia. Primero calcularemos el punto central de la circunferencia de intersección. Para ello calcularemos la ecuación de la recta que une los dos centros de las dos esferas y una vez calculada la intersecamos con el plano de intersección entre las dos esferas. La recta que une los dos centros está compuesta por los segmentos (2) y (3) de la Imagen 15, siendo (1), el punto de intersección con el plano (4), que a su vez es el plano de intersección entre las dos esferas.

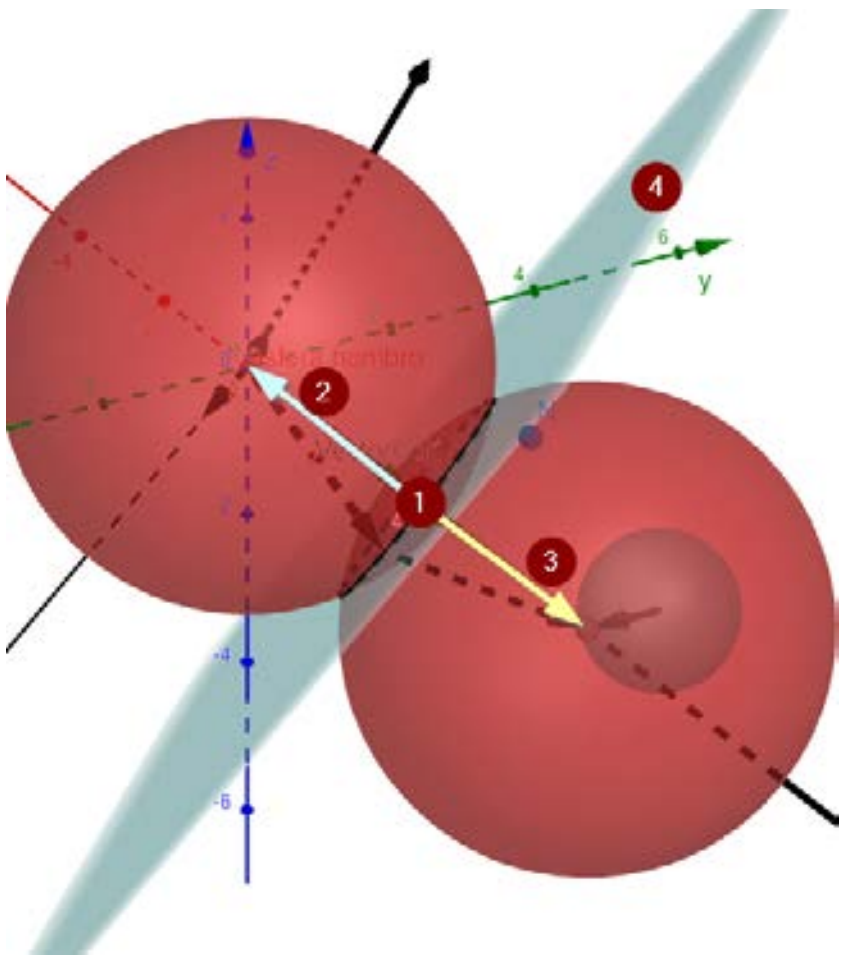

*Imagen 15. Cálculo del centro de la circunferencia de intersección de las dos esferas.*

El segmento que une ambos centros de ambas esferas comienza en $(0,0,0)$ y finaliza en $(a,b,c)$. Dado que el punto inicial es el origen de coordenadas, el vector que define la recta que parte en el origen de coordenadas y finaliza en el centro de la segunda esfera es:

$$\vec{d} = (a,b,c) \quad (28)$$

Siendo la ecuación paramétrica de la recta en el espacio

$$(x,y,z) = (0,0,0) + \beta(a,b,c) \quad (29)$$

$$\begin{cases} x = \beta a \\ y = \beta b \\ z = \beta c \end{cases} \quad (30)$$

Sustituimos el resultado de la ecuación $(30)$ en la ecuación $(18)$ y obtenemos:

$$\beta = \frac{{d_1}^2 - {d_2}^2 + a^2 + b^2 + c^2}{(2a^2 + 2b^2 + 2c^2)} = \frac{1}{2} \quad (31)$$

Al aplicar el resultado a la ecuación paramétrica de la recta (30) obtenemos el punto medio de la circunferencia de intersección (Ecuación (33)):

$$\begin{cases} m_x = \beta a = {}^3/_2 \\ m_y = \beta b = {}^3/_2 \\ m_z = \beta c = -{}^3/_2 \end{cases} \quad (32)$$

$$M = (m_x, m_y, m_z) = (\frac{3}{2}, \frac{3}{2}, -\frac{3}{2}) \quad (33)$$

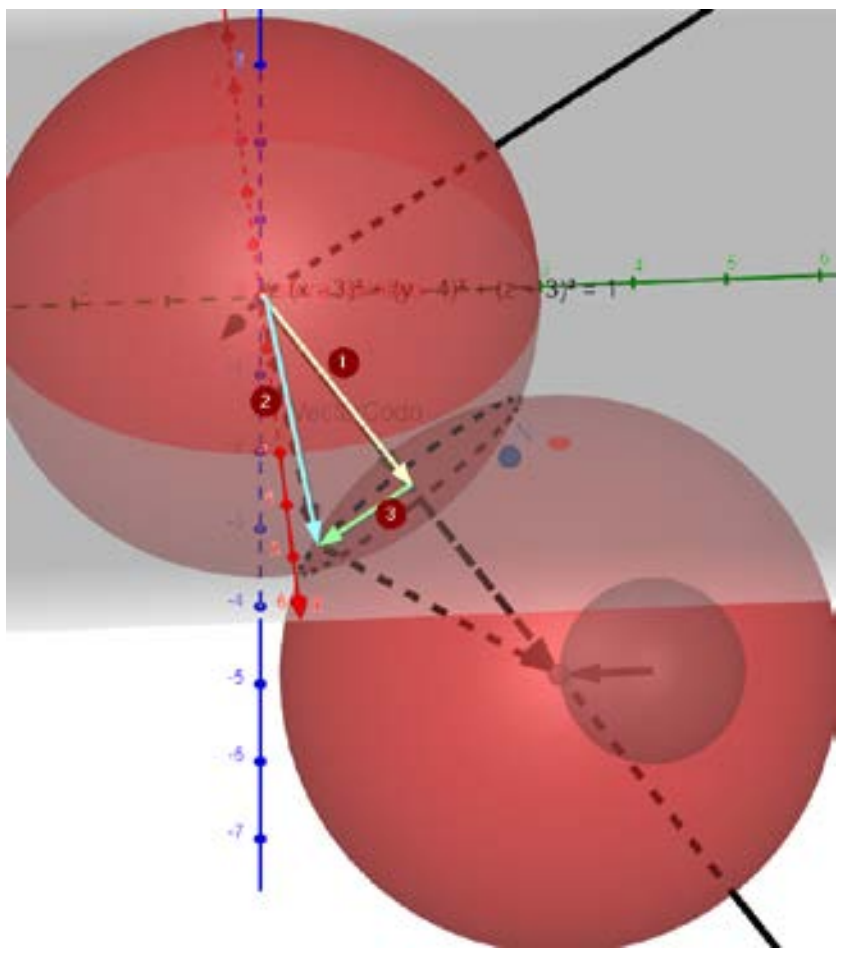

*Imagen 16. Radio de la circunferencia de intersección de las dos esferas.*

Para pintar una circunferencia necesitamos conocer el centro y el radio. El centro lo hemos calculado en con las ecuaciones (29) a (33). Dado que conocemos la longitud del brazo (segmento azul (2) de la Imagen 16) y conocemos la longitud del segmento que une el origen de coordenadas y el centro de la circunferencia de intersección (Ecuación 34)(segmento amarillo (1) de la Imagen 16), aplicando el teorema de Pitágoras calculamos el radio que está identificado con el segmento verde (3) de la Imagen 16) (ecuación (35)).

$$d = \sqrt{{m_x}^2 + {m_y}^2 + {m_z}^2} = 2.6 \quad (34)$$

$$r3 = \sqrt{{d_1}^2 - d^2} = \sqrt{3^2 - 2.6^2} = 1.5 \quad (35)$$

$$\vec{M} = \left(\frac{3}{2}, \frac{3}{2}, -\frac{3}{2}\right) \rightarrow Vector\ (1)\ de\ la\ \text{Imagen 16} \quad (36)$$

Una vez calculados los vectores que formarán el eje de coordenadas del plano $\overrightarrow{an_1}$, ecuación (25) y $\overrightarrow{bn_1}$, ecuación (27) y el centro de la circunferencia de intersección (ecuación (33)) podemos desarrollar la ecuación paramétrica tridimensional en la ecuación (34).

$$\vec{M} + r3 \cdot \overrightarrow{an_1} \cdot \cos(t) + r3 \cdot \overrightarrow{bn_1} \cdot \mathrm{sen}(t), 0 \le t \le 2\pi \quad (37)$$

Cualquier punto dentro de la circunferencia de la ecuación (34) es un valor válido para posicionar el codo, siempre que no tengamos restricciones físicas. En el caso de calcular la posición del brazo derecho, si el punto final que queremos alcanzar está pegado al cuerpo, el rango de posiciones se restringe. En la Imagen 17 podemos ver el centro del eje de coordenadas. El codo no puede posicionarse a la derecha (visto de frente) del eje vertical, dado que colisionaría con el cuerpo, con lo que podemos apreciar como la geometría del robot nos limita las posiciones válidas dentro de la circunferencia de intersección de las esferas para la posición del codo.

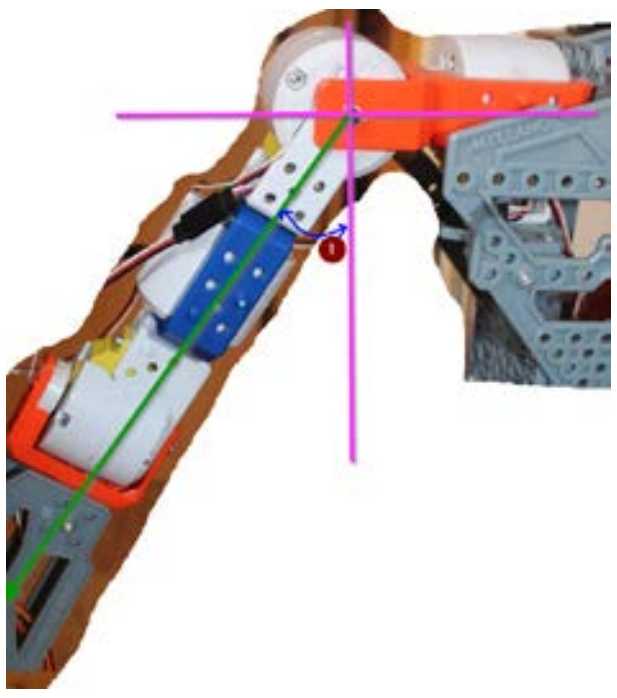

Imagen 17. Punto (0,0,0).

En un caso como el de la Imagen 17, el rango de posiciones válidas para el valor de $t$ en la ecuación (37) estaría entre π/2 y 3π/2 como muestra el arco de la Imagen 18, en donde se puedan ver marcados los puntos que corresponden a los valores 0 o 2π (1), π/2 (2), π (3) y 3π/2 (4). Si posicionamos el codo en el semicírculo no trazado quedaría delante del cuerpo y podría colisionar con él.

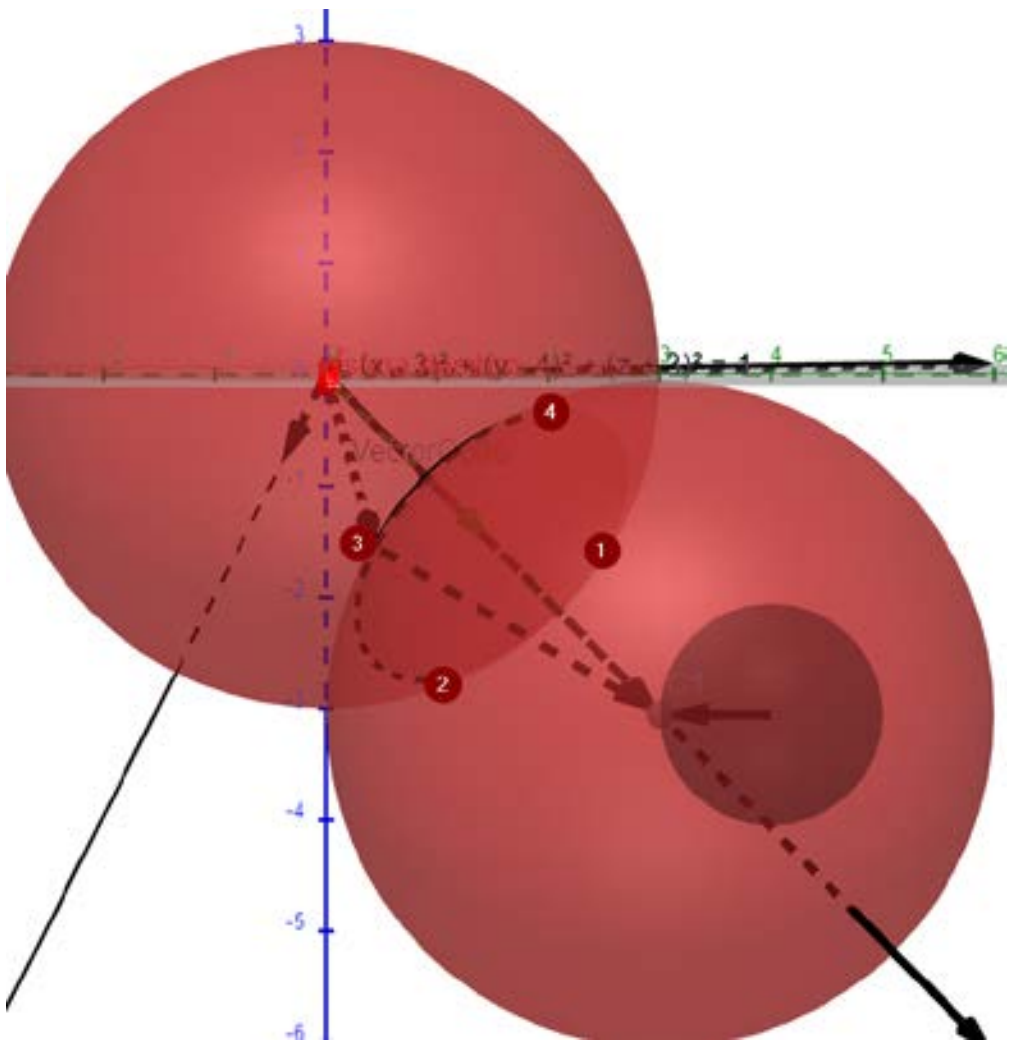

Imagen 18. Arco de posicionamiento válido para el codo.

Dentro de todas las soluciones válidas que son las que se encuentran dentro del arco definido por las posiciones físicamente posibles debemos seleccionar una de ellas. La seleccionada dependerá de la posición de partida, dado que elegiremos la que nos sitúe en una posición que sea alcanzable con el mínimo gasto de energía. Una vez seleccionada la posición del codo, debemos calcular los ángulos que deben alcanzar todos los servos para posicionar la articulación en la posición seleccionada.

Para seguir con nuestro ejemplo escogeremos la posición con el ángulo π (punto (3) de la Imagen 18) dentro de nuestra circunferencia de intersección, y a partir de esta posición del codo calcularemos todos los ángulos.

Las coordenadas del punto que representa el codo se calculan por las ecuaciones (38) a (40).

$$codo_x = M_x + r3 \cdot \cos(\pi) \cdot \overrightarrow{an_1} \cdot (1,0,0) + r3 \cdot \sin(\pi) \cdot \overrightarrow{bn_1} \cdot (1,0,0) \approx 2.56 \quad (38)$$

$$codo_y = M_y + r3 \cdot \cos(\pi) \cdot \overrightarrow{an_1} \cdot (0,1,0) + r3 \cdot \sin(\pi) \cdot \overrightarrow{bn_1} \cdot (0,1,0) \approx 0.44 \quad (39)$$

$$codo_z = M_z + r3 \cdot \cos(\pi) \cdot \overrightarrow{an_1} \cdot (0,0,1) + r3 \cdot \sin(\pi) \cdot \overrightarrow{bn_1} \cdot (0,0,1) = -\frac{3}{2} \quad (40)$$

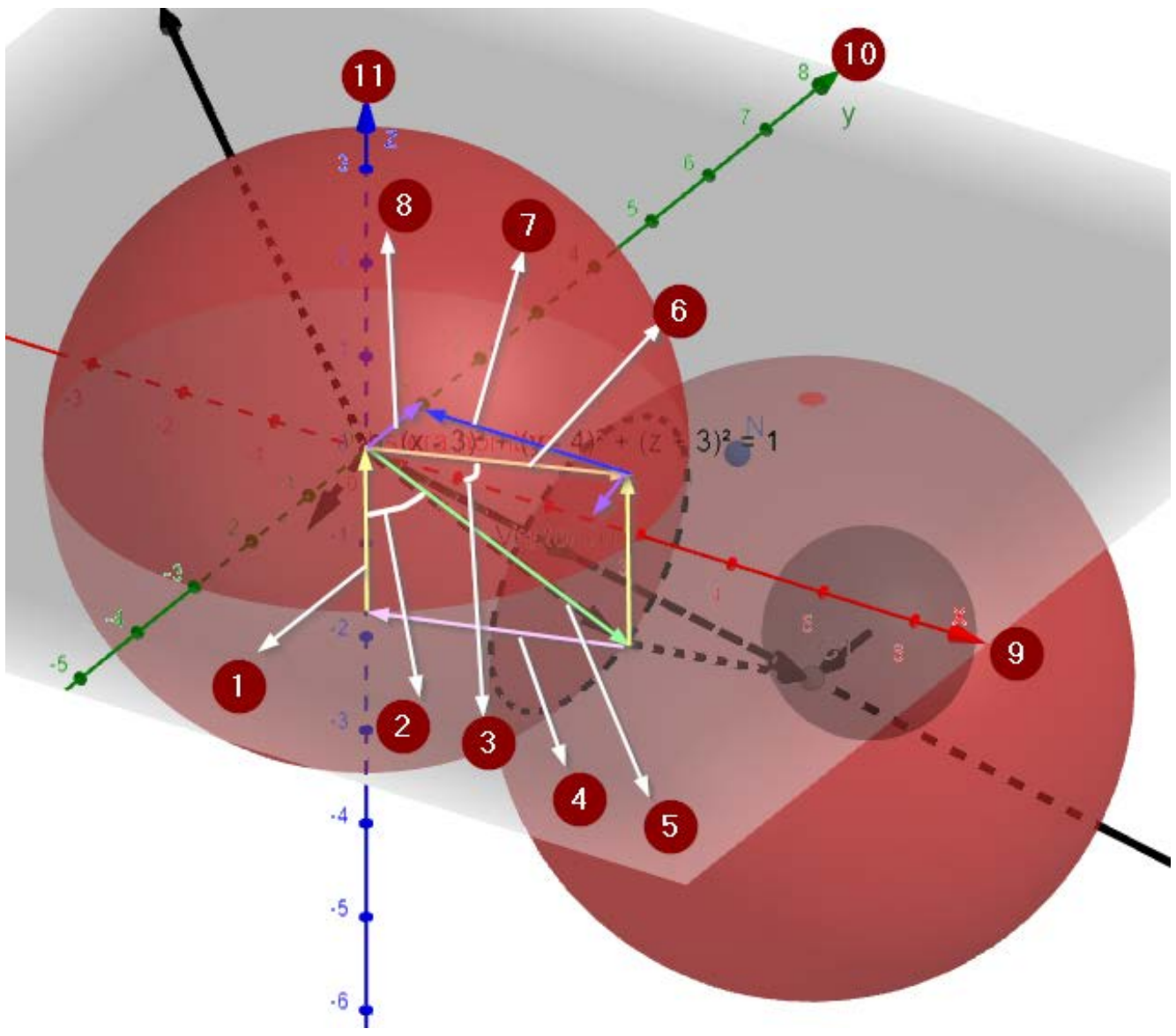

*Imagen 19. Imagen con los ángulos a resolver.*

En la Imagen 19 podemos apreciar todos los segmentos que entran en la resolución de la cinemática inversa de los dos ángulos del hombro. Los ejes están identificados con (9) el eje $X$, (10) el eje $Y$, y (11) el eje $Z$.En verde podemos ver el segmento que representa el brazo que parte del hombro y termina en el codo (5). Para resolver el ángulo (2) tenemos que disponer de la longitud de todos los segmentos del triángulo formado por los segmentos (1), (4), (5). En este caso el segmento (5) es la longitud del brazo y el segmento (1) es la posición sobre el eje Z del codo. Dado que los tres segmentos forman un triángulo rectángulo, aplicando el teorema de Pitágoras podemos resolver el segmento (4) como indica la ecuación (41). El valor (5) de la longitud del brazo procede de la ecuación (14).

$$catCodo_z = \sqrt{{d_1}^2 - {codo_z}^2} = \sqrt{3^2 - \left(-\frac{3}{2}\right)^2} \approx 2.6 \quad (41)$$

y una vez obtenidos todos los lados procedemos a resolver el ángulo (2) como indica la ecuación (42):

$$angHombro_z = \sin^{-1}\left(\frac{catCodo_z}{d_1}\right) = \sin^{-1}\left(\frac{2.6}{3}\right) \approx 1.05 \approx 60.16° \quad (42)$$

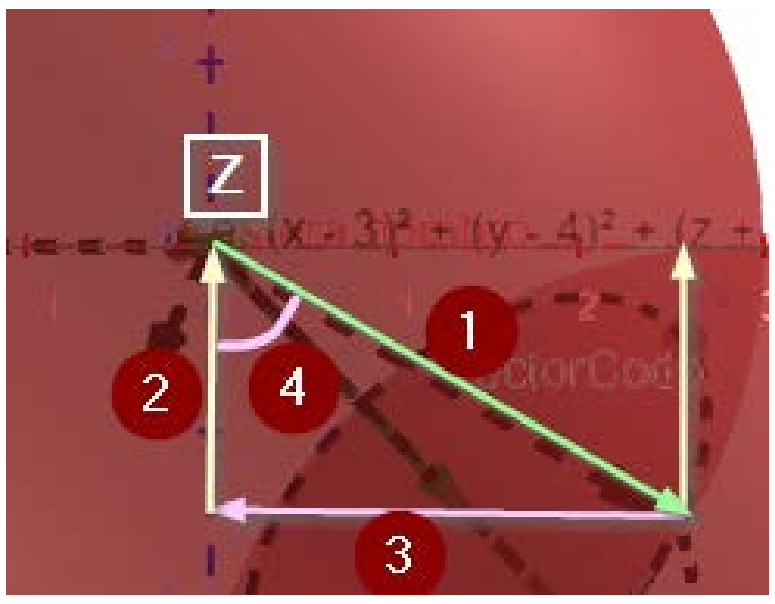

*Imagen 20. Vista lateral del segmento del hombro.*

Desde el eje Z conocemos la longitud del brazo (vector verde (1) en la Imagen 20) y la posición en el eje Z del codo (inicio vector amarillo (2)), pero nos falta el segmento (3) para poseer toda la información del triángulo rectángulo y calcular el ángulo vertical desde el eje Z (4). En la ecuación (41) se resuelve el segmento (3) y en la ecuación (42) se resuelve el ángulo (4).

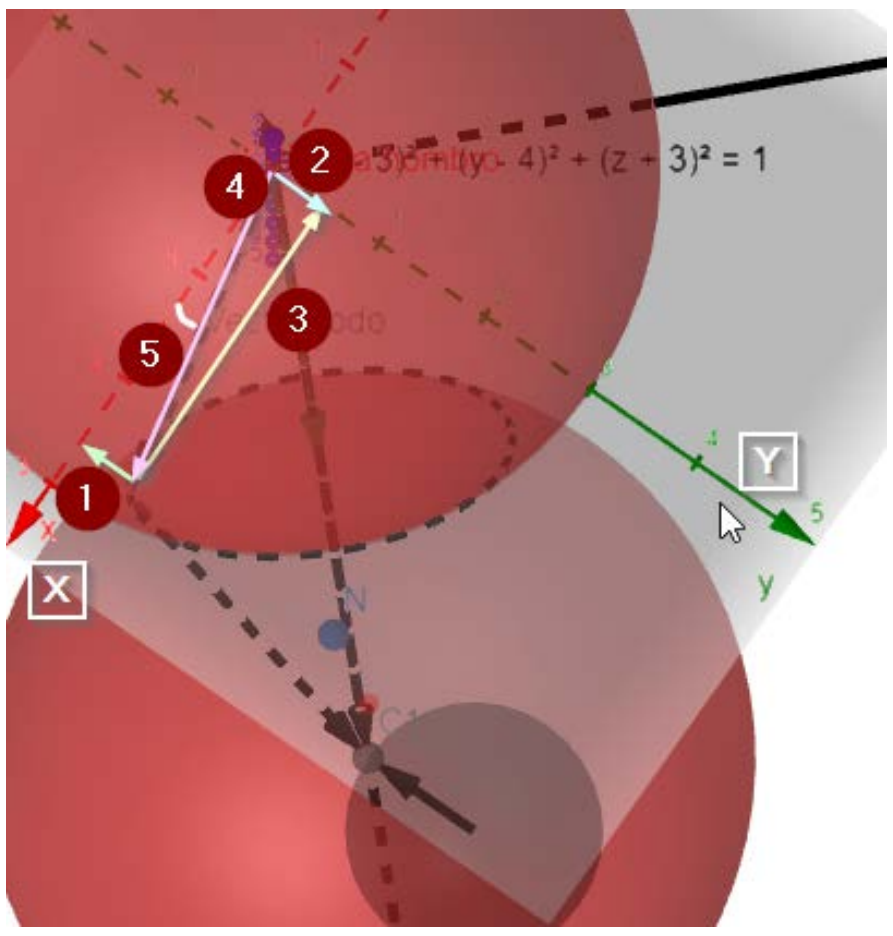

*Imagen 21. Resolución del ángulo del hombro sobre el eje XY.*

Una vez resuelto el ángulo sobre el eje Z, en el siguiente paso se resolverá el ángulo del hombro con respecto al eje X sobre el plano XY. En la Imagen 21 podemos ver la proyección del segmento del hombro sobre el plano XY. Debe calcularse el ángulo (5) que forma el eje X de coordenadas y el segmento morado (4) que es la proyección del vector del hombro en el plano XY. En la Imagen 21 se conoce el segmento (2) que es la posición Y del codo, pero no se conoce el segmento morado (4) que es la proyección del hombro sobre el plano XY y que será necesario para calcular el ángulo (5) que se obtiene en la ecuación (43).

$$hipBrazo_{xy} = \sqrt{{codo_x}^2 + {codo_y}^2} = \sqrt{2.56^2 + (0.44)^2} \approx 2.59 \qquad (43)$$

Finalmente se calcula el ángulo (5) de la Imagen 21 como indica la ecuación (44).

$$angHombro_x = cos^{-1}\left(\frac{codo_x}{hipBrazo_{xy}}\right) = cos^{-1}\left(\frac{2.56}{2.59}\right) \approx 0.15 \approx 8.59° \qquad (44)$$

A partir de este punto se deberán calcular los dos ángulos que deben aplicarse a los dos servos del codo para posicionar el antebrazo.

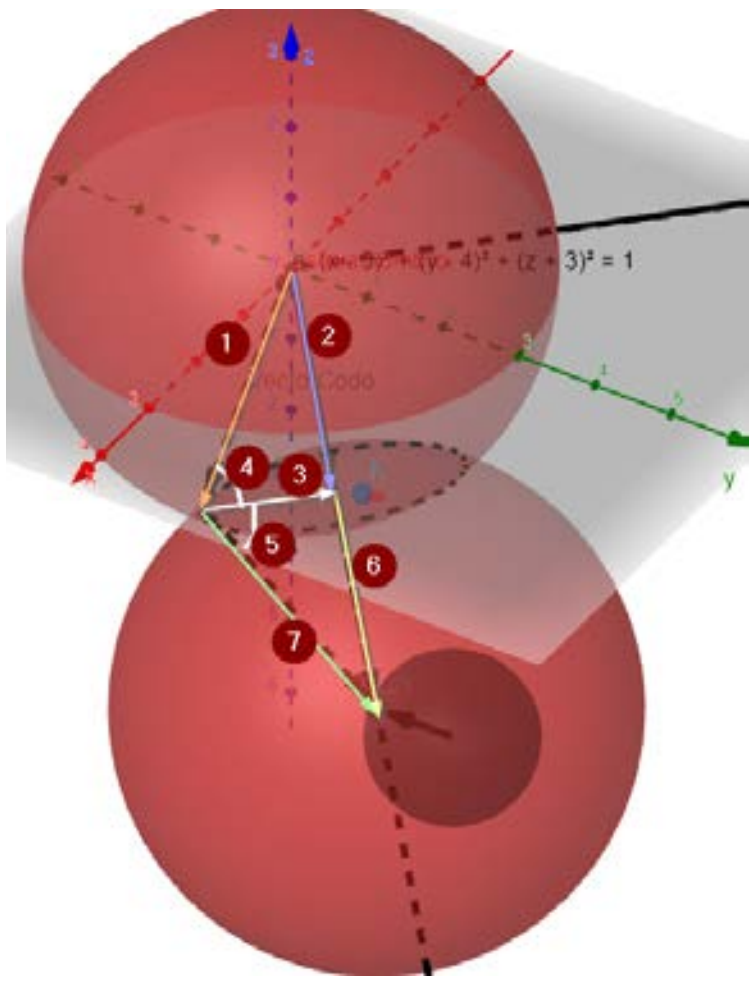

*Imagen 22. Ángulo frontal del codo.*

Para calcular el ángulo frontal del codo debe calcularse tanto el ángulo (4) como el ángulo (5) de la Imagen 22. Para resolver el ángulo (5) se necesita el segmento $\boldsymbol{catAntebrazo}$ (6) que une el centro de la circunferencia de intersección entre las dos esferas (ecuación (32)) y el punto de la muñeca (ecuación (10)), y el segmento $\boldsymbol{d_2}$ (7) que es la longitud del antebrazo (ecuación (15)).

$$\begin{aligned} catAntebrazo &= Distancia\left((M_x, M_y, M_z), (m_x, m_y, m_z)\right) \\ &= \sqrt{(M_x - m_x)^2 + \left(M_y - m_y\right)^2 + (M_z - m_z)^2} \\ &= \sqrt{\left(2 - \frac{3}{2}\right)^2 + \left(3 - \frac{3}{2}\right)^2 + \left(-3 - \left(-\frac{3}{2}\right)\right)^2} \approx 2.17 \end{aligned} \tag{45}$$

$$angCodo_2 = \sin^{-1}\left(\frac{catAntebrazo}{d_2}\right) = \sin^{-1}\left(\frac{2.17}{3}\right) \approx 0.81 = 45.41° \tag{46}$$

Para calcular el ángulo (4) necesitamos el segmento (2) que une el origen de coordenadas con el punto medio de la circunferencia de intersección de las esferas (ecuación (32)) y el segmento (1) que es la longitud del brazo $\boldsymbol{d_1}$ de la ecuación (14).

$$\begin{aligned} catBrazo &= Distancia\left((m_x, m_y, m_z), (0,0,0)\right) \\ &= \sqrt{(0 - m_x)^2 + \left(0 - m_y\right)^2 + (0 - m_z)^2} \approx 2.59 \end{aligned} \tag{47}$$

$$angCodo_1 = \sin^{-1}\left(\frac{catBrazo}{d_1}\right) = \sin^{-1}\left(\frac{2.59}{3}\right) \approx 1.05 = 60.16° \tag{48}$$

$$angCodo = andCodo_1 + andCodo_2 = 45.41 + 60.16 = 105.57° \tag{49}$$

Hasta este punto se ha calculado el ángulo frontal del codo y quedaría pendiente por calcular el ángulo lateral, ero este no se calcula, sino que se selecciona para quedarnos con una de las soluciones posibles dentro de la circunferencia de intersección. En el ejemplo actual, el ángulo seleccionado es π y fue seleccionado en para resolver las ecuaciones (38) a (40).

El último pasó será calcular los ángulos de la muñeca y una vez calculados, en caso de que no sea posible aplicarlos por las limitaciones físicas del robot, se deberá recalcular el punto de la muñeca en base a los ángulos posibles.

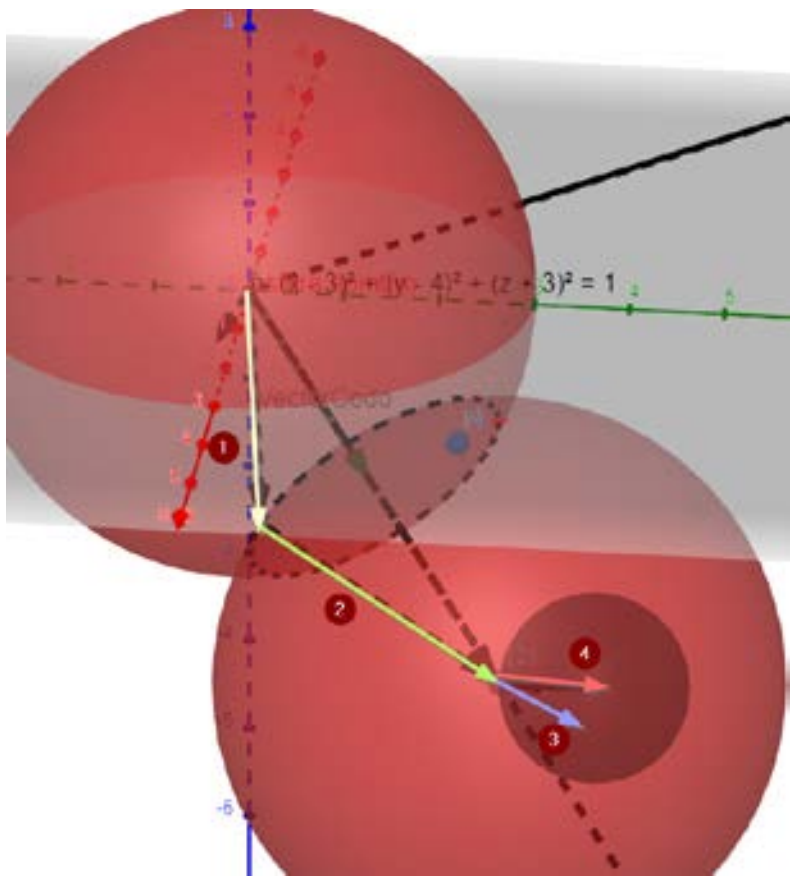

*Imagen 23. Ajuste del vector de la mano.*

Como se ve en la Imagen 23, inicialmente el vector de la mano (3) continúa con la misma dirección que el vector del antebrazo (2). En este punto estará girado el mismo ángulo que el codo dentro de la circunferencia de intersección de las dos esferas, por lo que conocemos el ángulo de rotación inicial. Nuestro objetivo es calcular los ángulos de rotación de la muñeca y mano para que el vector de la mano pase de su posición actual (3) hasta solaparse con la posición deseada (4).

El primer paso será conseguir que el plazo de desplazamiento de la mano sea el mismo que el plano que contiene los vectores (3) y (4). Este paso debe conseguirse rotando el servo (1) de la muñeca como se muestra en la Imagen 24. Una vez que los dos vectores se encuentren en el mismo plano, se empleará el servo (2) de la Imagen 24 para desplazar la mano de la posición indicada por el vector (3) de la Imagen 23 a la posición indicada por el vector (4).

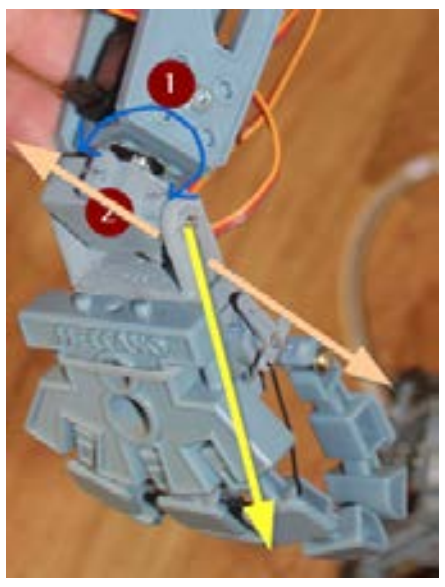

*Imagen 24. Servos de la muñeca.*

Debemos de tener en cuenta que en este punto conocemos la posición de la muñeca y conocemos la posición del extremo de la mano, con lo que sabemos con precisión la dirección que tiene que tomar la mano. Como vemos en la Imagen 24 disponemos dos servos para conseguir que la mano tenga la misma dirección que el vector de la mano calculado geométricamente. El servo (1) que nos permite girar la muñeca y el servo (2) que nos permite mover la mano. Dado que conocemos el vector del antebrazo y el vector de la mano, sabemos que con el servo (2) tenemos que rotar el ángulo que hay entre el vector del antebrazo y el vector de la mano, pero no podemos aplicar esta rotación hasta que el plano de esta rotación contenga el vector de la mano. Por este motivo, primero calcularemos el plano de rotación actual de la mano. Una vez calculado, calcularemos el ángulo entre este plano y el vector de la mano, rotaremos la muñeca hasta conseguir que el ángulo entre el vector del plano y el vector de la mano sea 0 y finalmente rotaremos la mano según el ángulo entre el vector del antebrazo y el vector de la mano.

El primer paso será calcular el plano de desplazamiento del servo (2) de la Imagen 24. Para ello emplearemos los tres puntos de referencia de la Imagen 25. El plano obtenido será perpendicular al desplazamiento de mano, teniendo en cuenta que la posición inicial es la que podemos ver en la Imagen 2.

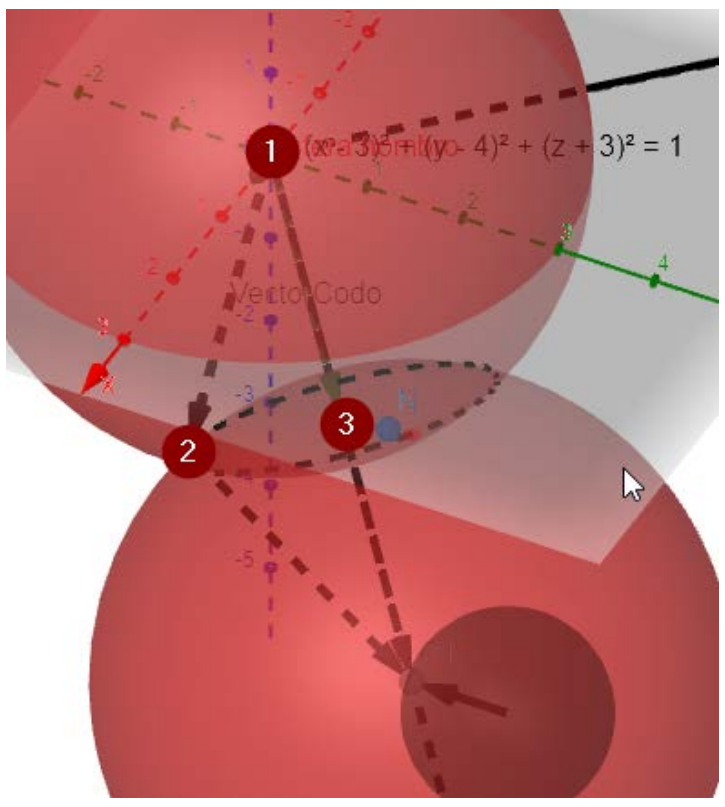

*Imagen 25. Puntos de cálculo del plano perpendicular al desplazamiento de la mano.*

$$P_1 = (0,0,0), P_2 = \left(codo_x, codo_y, codo_z\right), P_3 = (m_x, m_y, m_z) \qquad (50)$$

$$\overrightarrow{P_1P_2} = \left(codo_x - 0, codo_y - 0, codo_z - 0\right) = (2.56, 0.44, -1.5) \qquad (51)$$

$$\overrightarrow{P_1P_3} = \left(m_x - 0, m_y - 0, m_z - 0\right) = (1.5, 1.5, -1.5) \qquad (52)$$

$$\overrightarrow{P_1P_2} \times \overrightarrow{P_1P_3} = \begin{vmatrix} i & j & k \\ 2.56 & 0.44 & -1.5 \\ 1.5 & 1.5 & -1.5 \end{vmatrix} \qquad (53)$$

$$+\begin{vmatrix} 0.44 & -1.5 \\ 1.5 & -1.5 \end{vmatrix} i - \begin{vmatrix} 2.56 & -1.5 \\ 1.5 & -1.5 \end{vmatrix} i + \begin{vmatrix} 2.56 & 0.44 \\ 1.5 & 1.5 \end{vmatrix} k \qquad (54)$$

$$\left(-1.5 \cdot 0.44 - (-1.5 \cdot 1.5)\right)i - \left((-1.5 \cdot 2.56) - (-1.5 \cdot 1.5)\right)j + ((1.5 \cdot 2.56) - (1.5 \cdot 0.44))k \qquad (55)$$

$$(-0.66 + 2.25)i - (-3.84 + 2.25)j + (3.84 - 0.66)k \qquad (56)$$

$$1.59i + 1.59j + 3.18k \rightarrow Ecuación\ del\ plano \qquad (57)$$

$$\vec{N} = (1.59, 1.59, 3.18) \rightarrow Vector\ perpendicular\ al\ plano \qquad (58)$$

$$\vec{N} \cdot [P - P_1] = 0 \rightarrow Ecuación\ del\ plano\ que\ pasa\ por\ el\ punto\ P_1 \qquad (59)$$

$$(1.59, 1.59, 3.18) \cdot [(x, y, z) - (0,0,0)] = 0 \quad (60)$$

$$(1.59, 1.59, 3.18) \cdot (x, y, z) = 0 \quad (61)$$

$$1.59x + 1.59y + 3.18z = 0 \quad (62)$$

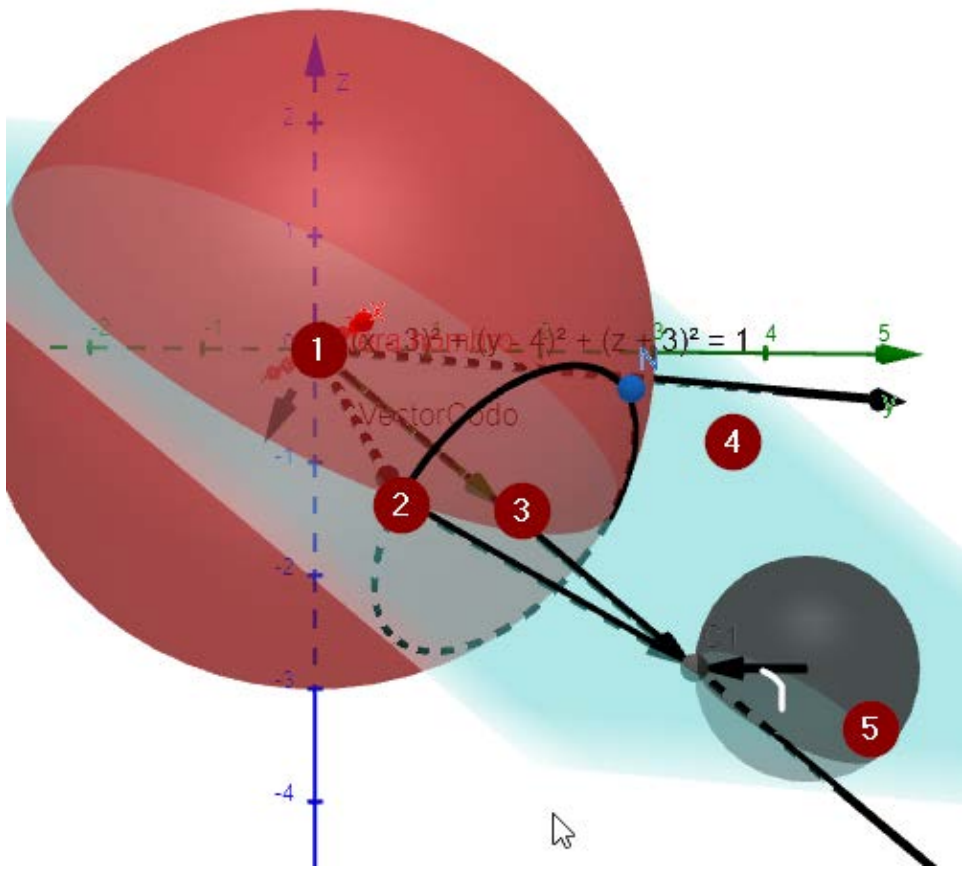

*Imagen 26. Ángulo del plano de desplazamiento de la mano con el vector de la mano.*

El servo (2) de la Imagen 24 se desplaza por el plano (4) de la Imagen 26. Esto implica que si el punto final de la mano no se encuentra en el plano (4), aunque rotemos el servo no será posible alcanzar el punto final, pero dado que conocemos el punto de la muñeca, conocemos el punto final y conocemos el plano de desplazamiento de la mano, lo único que debemos hacer es conseguir que el plano de desplazamiento de la mano rote para que llegue a contener el vector que une el punto de la muñeca y el punto final. Una vez rotado calcularemos el ángulo entre el antebrazo y la mano y podremos rotar el servo (2) de la Imagen 24 para que la mano del robot y el vector de la mano en nuestra representación geométrica se superpongan.

Una vez que tenemos el plano de rotación de la mano, hay que obtener la ecuación de la recta que pasa por el punto de la muñeca y el punto del extremo de la mano.

El siguiente paso será obtener el ángulo entre el plano y la recta para ello calcularemos el vector director de la recta y el vector perpendicular al plano.

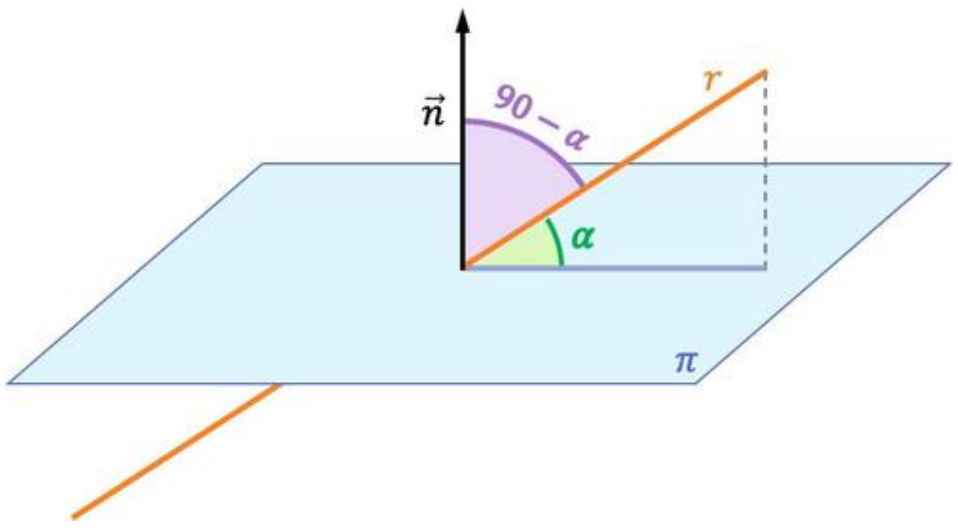

*Imagen 27. Ángulo de la recta r con el plano (GeometriaAnalitica.com).*

$$\cos(90° - \alpha) = \frac{|\overrightarrow{d_r} - \overrightarrow{n_\pi}|}{|\overrightarrow{d_r}| \cdot |\overrightarrow{n_\pi}|} \tag{63}$$

$$\vec{r} = P + t\vec{v} \tag{64}$$

$$A = (P_x, P_y, P_z), B = (a, b, c) \tag{65}$$

$$P = A \tag{66}$$

$$\vec{v} = \overrightarrow{AB} = (a, b, c) - (P_x, P_y, P_z) = (3,3,-3) - (3,4,-3) = (0,-1,0) \tag{67}$$

$$\vec{r} = (3,3,-3) + t(0,-1,0) \tag{68}$$

$$(x, y, z) = (3,3,-3) + t(0,-1,0) \tag{69}$$

$$(x, y, z) = (3,3,-3) + (0, -t, 0) \tag{70}$$

$$(x, y, z) = (3 + 0, 3 - t, -3 + 0) \tag{71}$$

$$\begin{cases} x = 3 \\ y = 3 - t \\ z = -3 \end{cases} \tag{72}$$

$$\overrightarrow{n_\pi} = (1.59, 1.59, 3.18) \rightarrow ecuación\ (58) \tag{73}$$

$$\overrightarrow{d_r} = (0,-1,0) \tag{74}$$

$$\cos(90° - \alpha) = \frac{|\overrightarrow{d_r} \cdot \overrightarrow{n_\pi}|}{|\overrightarrow{d_r}| \cdot |\overrightarrow{n_\pi}|} \tag{75}$$

$$\cos(90° - \alpha) = \frac{|(1.59 \cdot 0 + 1.59 \cdot (-1) + 3.18 \cdot 0)|}{\sqrt{0^2 + (-1)^2 + 0^2} \cdot \sqrt{1.59^2 + 1.59^2 + 3.18^2}} = \frac{1.59}{3.89} \approx 0.41 \tag{76}$$

$$90° - \alpha = \cos^{-1}(0.41) \tag{77}$$

$$\alpha = 90° - 65.89° = 24.11° \tag{78}$$

Como sabemos que el plano de rotación estaba desplazado 90° grados respecto a la posición inicial del mismo, y dado que entre el plano y el vector de la mano hay un ángulo de 24.11°, la rotación total del servo de la muñeca debe ser de 90°+24.11° = 114.11°.

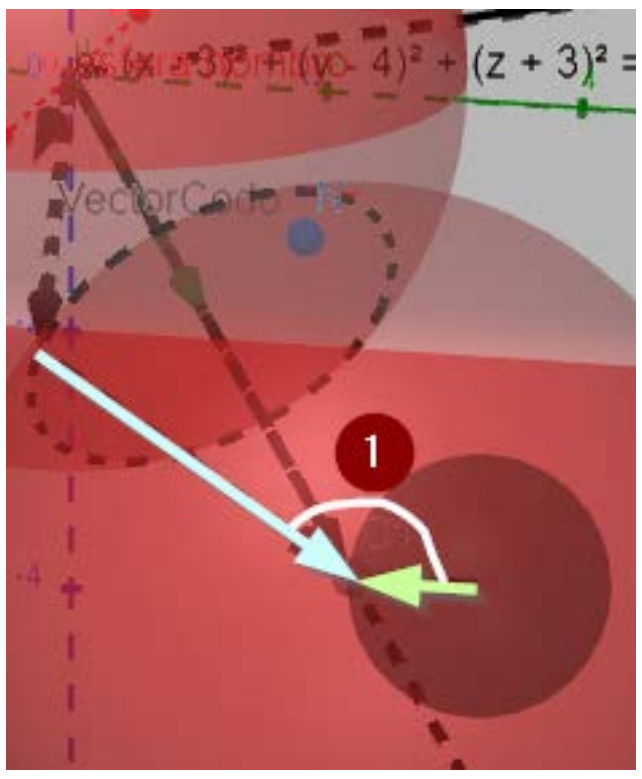

*Imagen 28. Ángulo final de rotación de la mano.*

Una vez alineado el plano de rotación de la mano con el vector de la mano en la representación geométrica, el último paso será calcular el ángulo entre el vector del antebrazo (azul) y el vector de la mano (verde) para desplazar la mano a la posición final (ángulo (1) de la Imagen 28).

$$\cos(\alpha) = \frac{|\overrightarrow{v_1} \cdot \overrightarrow{v_2}|}{|\overrightarrow{v_1}| \cdot |\overrightarrow{v_2}|} \quad (79)$$

Desplazamos los dos vectores al origen de coordenadas

$$\vec{v}_1 = (a.b.c) - (codo_x, codo_y, codo_z) \quad (80)$$

$$\vec{v}_2 = (a.b.c) - (P_x, P_y, P_z) \quad (81)$$

Si restamos las componentes de ambos puntos para conseguir que los segmentos se inicien en el origen de coordenadas.

$$\vec{v}_1 = (0,0,0) - (codo_x - a, codo_y - b, codo_z - c) = (2.56 - 3, 0.44 - 3, -1.5 - (-3)) = (-0.44, -2.56, 1.5) \quad (82)$$

$$\vec{v}_2 = (0,0,0) - (P_x - a, P_y - b, P_z - c) = (3 - 3, 4 - 3, -3 - (-3)) = (0,1,0) \quad (83)$$

Ahora podemos calcular su ángulo

$$\cos(\alpha) = \frac{|\overrightarrow{v_1} \cdot \overrightarrow{v_2}|}{|\overrightarrow{v_1}| \cdot |\overrightarrow{v_2}|} \quad (84)$$

$$\cos(\alpha) = \frac{|(-0.44, -2.56, 1.5) \cdot (0,1,0)|}{\left|\sqrt{(-0.44)^2 + (-2.56)^2 + (1.5)^2}\right| \cdot \left|\sqrt{(0)^2 + (1)^2 + (0)^2}\right|} = \frac{|(-0.44) \cdot 0 + (-2.56) \cdot 1 + 1.5 \cdot 0)|}{3 \cdot 1} = \frac{-2.56}{3} \approx -0.85 \quad (85)$$

$$\alpha = \cos^{-1}(-0.85) \approx 2.59 = 148.40° \quad (86)$$

# 6. Referencias

[1] N. Koban, N. Çavli, H. Doğan, y E. Benli, «7 DOF Robotic Arm Inverse Kinematic Analysis with Cuckoo and Whale Algorithms», en *2023 Innovations in Intelligent Systems and Applications Conference (ASYU)*, oct. 2023, pp. 1-5. doi: 10.1109/ASYU58738.2023.10296820.

[2] L. Yang y G. Zhang, «An Analytic Solution Study for a 7-DOF Redundant Robot Manipulator», en *2018 IEEE 8th Annual International Conference on CYBER Technology in Automation, Control, and Intelligent Systems (CYBER)*, jul. 2018, pp. 1057-1061. doi: 10.1109/CYBER.2018.8688249.

[3] C. Chen *et al.*, «Analysis and simulation of kinematics of 5-DOF nuclear power station robot manipulator», en *2014 IEEE International Conference on Robotics and Biomimetics (ROBIO 2014)*, dic. 2014, pp. 2025-2030. doi: 10.1109/ROBIO.2014.7090634.

[4] Oscar Ramirez, «Cinemática Directa: Parámetros de Denavit Hartenberg - YouTube». Accedido: 13 de julio de 2024. [En línea]. Disponible en: https://www.youtube.com/watch?v=V_IIeLJzR44

[5] A. A. Al-Hamadani y M. Z. Al-Faiz, «Design and Implementation of Inverse Kinematics Algorithm to Manipulate 5-DOF Humanoid Robotic Arm», en *2021 International Conference on Innovation and Intelligence for Informatics, Computing, and Technologies (3ICT)*, sep. 2021, pp. 693-697. doi: 10.1109/3ICT53449.2021.9581570.

[6] K. V. Shihabudheen y G. N. Pillai, «Evolutionary fuzzy extreme learning machine for inverse kinematic modeling of robotic arms», en *2015 39th National Systems Conference (NSC)*, dic. 2015, pp. 1-6. doi: 10.1109/NATSYS.2015.7489105.

[7] R. R. Kumar y P. Chand, «Inverse kinematics solution for trajectory tracking using artificial neural networks for SCORBOT ER-4u», en *2015 6th International Conference on Automation, Robotics and Applications (ICARA)*, feb. 2015, pp. 364-369. doi: 10.1109/ICARA.2015.7081175.

[8] A. Sriram, A. R. R, R. Krishnan, S. Jagadeesh, y K. Gnanasekaran, «IoT-Enabled 6DOF Robotic Arm with Inverse Kinematic Control: Design and Implementation», en *2023 IEEE World Conference on Applied Intelligence and Computing (AIC)*, jul. 2023, pp. 795-800. doi: 10.1109/AIC57670.2023.10263943.

[9] X. Zheng, Y. Zheng, Y. Shuai, J. Yang, S. Yang, y Y. Tian, «Kinematics analysis and trajectory planning of 6-DOF robot», en *2019 IEEE 3rd Information Technology, Networking, Electronic and Automation Control Conference (ITNEC)*, mar. 2019, pp. 1749-1754. doi: 10.1109/ITNEC.2019.8729280.

[10] X. Zhang, M. Liu, L. Hu, S. Gou, y S. Wang, «Kinematics Analysis of a five-degree-of-freedom lightweight Explosive Ordnance Disposal robotic arm», en *2022 China Automation Congress (CAC)*, nov. 2022, pp. 2567-2571. doi: 10.1109/CAC57257.2022.10056037.

[11] «Robótica con Python - Cinemática inversa con Sympy - YouTube». Accedido: 13 de julio de 2024. [En línea]. Disponible en: https://www.youtube.com/watch?v=8qlfgI88nBk

[12] M. Abbas, J. Narayan, y S. K. Dwivedy, «Simulation Analysis for Trajectory Tracking Control of 5-DOFs Robotic Arm using ANFIS Approach», en *2019 5th International Conference On Computing, Communication, Control And Automation (ICCUBEA)*, sep. 2019, pp. 1-6. doi: 10.1109/ICCUBEA47591.2019.9128742.

[13] M. Crenganiş, R. Breaz, G. Racz, y O. Bologa, «The inverse kinematics solutions of a 7 DOF robotic arm using Fuzzy Logic», en *2012 7th IEEE Conference on Industrial Electronics and Applications (ICIEA)*, jul. 2012, pp. 518-523. doi: 10.1109/ICIEA.2012.6360783.

[14] S. Xu, «Trajectory planning of 6-DOF manipulator based on spatial ant colony algorithm and obstacle avoidance», en *2023 IEEE International Conference on Control, Electronics and Computer Technology (ICCECT)*, abr. 2023, pp. 18-22. doi: 10.1109/ICCECT57938.2023.10140314.